\documentclass{article}

\usepackage[preprint,nonatbib]{neurips_2026}

\usepackage[utf8]{inputenc} %
\usepackage[T1]{fontenc}    %
\usepackage{url}            %
\usepackage{booktabs}       %
\usepackage{amsfonts}       %
\usepackage{nicefrac}       %
\usepackage{microtype}      %
\usepackage{xcolor}         %
\usepackage{amsmath}

\definecolor{neuripsblue}{rgb}{0.21,0.49,0.74}

\usepackage{url}            %
\usepackage{booktabs}       %
\usepackage{amsfonts}       %
\usepackage{nicefrac}       %
\usepackage{microtype}      %
\usepackage{pifont}         %
\usepackage{makecell}
\usepackage{colortbl}
\usepackage{graphicx}
\usepackage{amsmath,amsfonts,amssymb,amsthm}
\usepackage{color}
\usepackage{caption}
\usepackage{subcaption}
\usepackage{cite}
\usepackage{xspace}
\usepackage{array}
\usepackage{tikzducks}
\usepackage{wrapfig}
\usepackage[titletoc,title]{appendix}
\usepackage{lscape}
\usepackage{enumitem}
\usepackage{multirow}
\usepackage{tcolorbox}
\usepackage{xspace}
\usepackage{pifont}
\usepackage{standalone}
\usepackage{tikz}
\usetikzlibrary{positioning, backgrounds, calc, arrows.meta, fit}

\usetikzlibrary{calc}
\usepackage{graphicx}

\usepackage{pgfplots}
\usepackage{pgfplotstable}
\pgfplotsset{compat=1.18} %
\usetikzlibrary{positioning}
\usepackage{sansmath}
\usetikzlibrary{arrows.meta}

\usepackage{amsmath, amssymb}
\usepackage{algorithm}
\usepackage{fontawesome5}
\usepackage{fdsymbol}

\usepackage{wrapfig}

\usepackage{bbold}
\tcbuselibrary{listings, breakable}

\usepackage{xcolor}
\definecolor{lightgray}{gray}{0.55}
\definecolor{origgray}{RGB}{120,120,120}
\definecolor{amethyst}{RGB}{153,102,204}
\definecolor{dar}{RGB}{255,140,0}
\definecolor{da}{RGB}{45,100,160}

\newcommand{\boldparagraph}[1]{\noindent{\bf #1}}

\def\xS{x_{\text{src}}}
\def\xT{x_{\text{trg}}}
\def\pStoT{p_{\text{src} \rightarrow \text{trg}}}

\makeatletter
\DeclareRobustCommand\onedot{\futurelet\@let@token\@onedot}
\def\@onedot{\ifx\@let@token.\else.\null\fi\xspace}
\def\eg{e.g\onedot} 
\def\ie{i.e\onedot} 
\def\cf{cf\onedot} 
\def\vs{vs\onedot}

\makeatother

\newcommand{\mstd}[2]{#1{\tiny\textcolor{gray}{\,$\pm$#2}}}

\newcommand{\datasetName}{V-CAPE\xspace}
\newcommand{\datasetNameAcronym}{Visual Consistency Assessment
for Product Image Editing\xspace}
\newcommand{\datasetNameReal}{V-CAPE-R\xspace}
\newcommand{\datasetNameSynthetic}{V-CAPE-S\xspace}

\newcommand{\figref}[1]{Fig.~\ref{#1}}
\newcommand{\secref}[1]{Section~\ref{#1}}
\newcommand{\eqnref}[1]{Eq.~\eqref{#1}}
\newcommand{\tabref}[1]{Table~\ref{#1}}

\usepackage[
    pagebackref=true,
    colorlinks=true,
    citecolor=neuripsblue,
    linkcolor=neuripsblue,
    urlcolor=neuripsblue
]{hyperref}

\title{V-CAPE: Visual Consistency Assessment \\ for Product Image Editing}

\author{
Iaroslav Melekhov$^\dagger$\quad Roberto Alcover-Couso$^\dagger$\quad Francesco Taioli\quad Daniel Coelho\quad \And Artur Bekasov$^{1}$\quad Virginia Fernandez Arguedas\quad Jose Migues Grande Saiz\vspace{8pt}\\
Amazon\qquad $^{1}$Faculty\vspace{8pt}\\
}

\begin{document}

\maketitle
\begingroup
\renewcommand{\thefootnote}{\dagger}
\footnotetext{Equal contribution. Email: \url{imlkhv@amazon.com}, \url{ralcover@amazon.com}}
\endgroup

\newcommand{\plotoverview}{
\begin{figure}[t!]
    \centering
    \includegraphics[width=\linewidth]{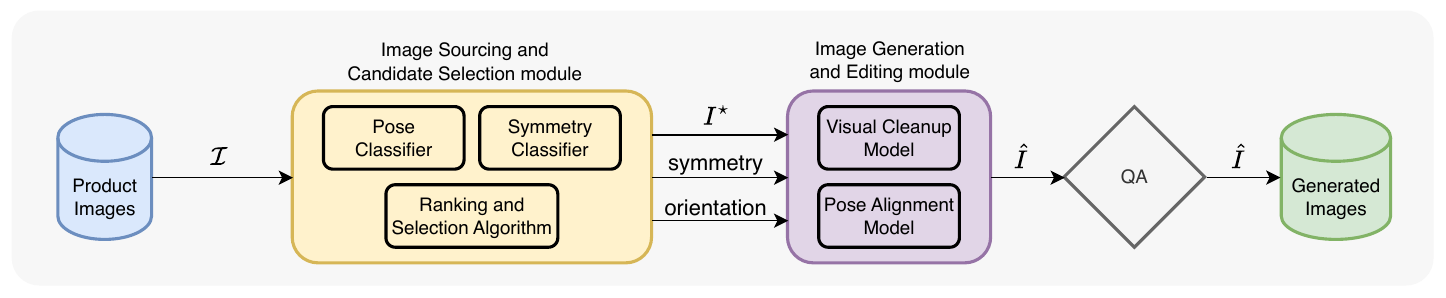}
    \caption{
        \textbf{\datasetNameReal generation pipeline.} For a given product type, the system aggregates all product images $\mathcal{I}$ and then ranks candidate images by compatibility with visual consistency requirements, and a single source image $I^\star$ is selected. The Image Generation and Editing module removes background clutter via a text-conditioned editing prompt and conditionally applies pose alignment based on predicted pose and symmetry. The edited image $\hat{I}$ is then validated by the QA pipeline.
    }
    \label{fig:overview_detailed}
\end{figure}
}

\newcommand{\vcapedatasetoverview}{
\begin{figure}[t!]
    \centering
    \includegraphics[width=\linewidth]{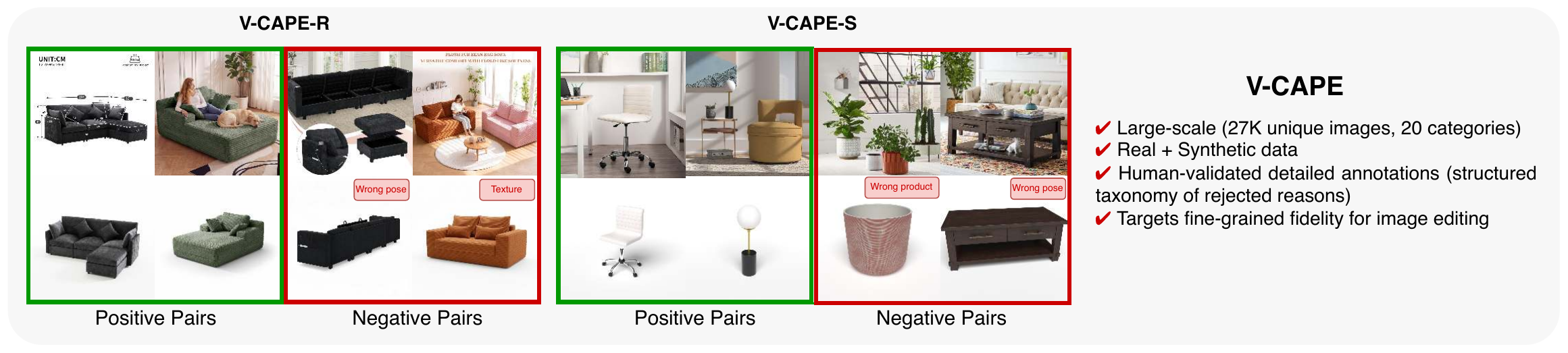}
    \caption{
        \textbf{\datasetName} is a large-scale dataset for automated quality assessment in generative image editing, targeting fine-grained visual fidelity. The dataset combines real-world generative outputs (\datasetNameReal) with controlled synthetic samples (\datasetNameSynthetic), comprising $27$K unique images ($40$K source--target pairs). Each sample is annotated by multiple trained annotators with a structured taxonomy of rejection reasons, going beyond binary labels to capture diverse failure modes.
    }
    \label{fig:dataset_overview}
\end{figure}
}
\newcommand{\tablefig}{
\begin{figure*}[t]
  \centering
  
  \pgfplotsset{
    myPlotStyle/.style={
      ybar, 
      bar width=11pt, 
      ymin=0, ymax=1.0,
      height=5.0cm,
      ymajorgrids, 
      grid style={gray!25},
      xtick=data, 
      enlarge x limits=0.22,
      tick label style={font=\footnotesize},
      title style={font=\small\bfseries},
      error bars/error bar style={line width=0.5pt, black},
      error bars/y dir=both, 
      error bars/y explicit,
    },
    myColors/.style={
      cycle list={
        {fill=blue!55!black!70,  fill opacity=0.6, draw=black!60},
        {fill=teal!70,           fill opacity=0.6, draw=black!60},
        {fill=orange!75,         fill opacity=0.6, draw=black!60},
      },
    }
  }

  \begin{tikzpicture}
    \begin{axis}[
      myPlotStyle, myColors,
      width=0.60\textwidth,
      title={\datasetNameSynthetic (3-class)}, 
      ylabel={Score},
      symbolic x coords={Acc,Front,Left,Right}
    ]
      \addplot+ coordinates {(Acc,0.382)+-(0,0.077) (Front,0.487)+-(0,0.098) (Left,0.352)+-(0,0.185) (Right,0.123)+-(0,0.106)}; 
      \addplot+ coordinates {(Acc,0.435)+-(0,0.101) (Front,0.478)+-(0,0.061) (Left,0.417)+-(0,0.145) (Right,0.377)+-(0,0.153)}; 
      \addplot+ coordinates {(Acc,0.588)+-(0,0.003) (Front,0.592)+-(0,0.020) (Left,0.658)+-(0,0.036) (Right,0.473)+-(0,0.081)}; 
    \end{axis}
  \end{tikzpicture}
  \hspace{0.05\textwidth}
  \begin{tikzpicture}
    \begin{axis}[
      myPlotStyle, myColors,
      width=0.44\textwidth,
      title={\datasetNameReal (2-class)},
      symbolic x coords={Acc,Front,Left}
    ]
      \addplot+ coordinates {(Acc,0.625)+-(0,0.085) (Front,0.133)+-(0,0.023) (Left,0.758)+-(0,0.067)}; 
      \addplot+ coordinates {(Acc,0.754)+-(0,0.005) (Front,0.457)+-(0,0.014) (Left,0.841)+-(0,0.003)}; 
      \addplot+ coordinates {(Acc,0.883)+-(0,0.040) (Front,0.568)+-(0,0.076) (Left,0.934)+-(0,0.024)}; 
    \end{axis}
  \end{tikzpicture}

  \vspace{-0.1cm}

  \begin{tikzpicture}
    \begin{axis}[
      hide axis, 
      xmin=1, xmax=2, ymin=1, ymax=2,
      scale only axis,
      width=1.0\linewidth, 
      height=1.0cm,
      myColors,
      legend style={
        draw=none, 
        fill=none, 
        font=\scriptsize,
        legend columns=-1, 
        at={(0.5,0.5)}, 
        anchor=center, 
        /tikz/every even column/.append style={column sep=30pt},
        legend image code/.code={
          \draw[draw=none, fill=##1] (0cm,-0.1cm) rectangle (0.5cm,0.1cm);
        }
      }
    ]
      \addplot+ coordinates {(0,0)}; 
      \addplot+ coordinates {(0,0)}; 
      \addplot+ coordinates {(0,0)}; 
      
      \legend{Contrastive vision-language models, Self-supervised vision encoders, Orientation estimation models}
    \end{axis}
  \end{tikzpicture}

  \vspace{-0.2cm}

  \caption{\textbf{Aggregated zero-shot orientation classification results.} Comparison of 
  model families across \datasetNameSynthetic (3-class) and \datasetNameReal (2-class). Bars 
  represent the mean performance within each technique family, while the error bars visualize the standard 
  deviation (std) across the family's individual models.}
  \label{fig:orient_merged}
\end{figure*}
}
\newcommand{\vllmfigurecomparison}{
\begin{figure}[t!]
    \centering
    \begin{subfigure}[b]{\linewidth}
    \includegraphics[width=\linewidth]{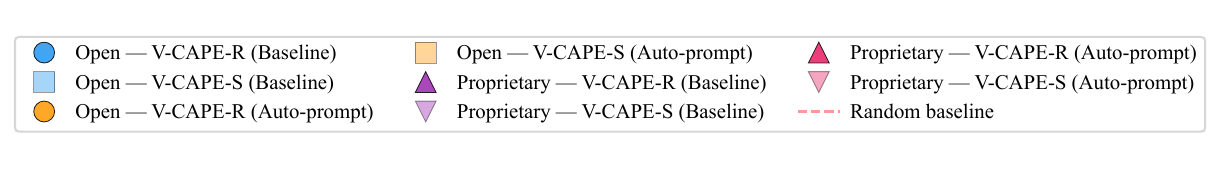}
    \end{subfigure}\hfill
    \begin{subfigure}[b]{0.5\linewidth}
    \includegraphics[width=\linewidth]{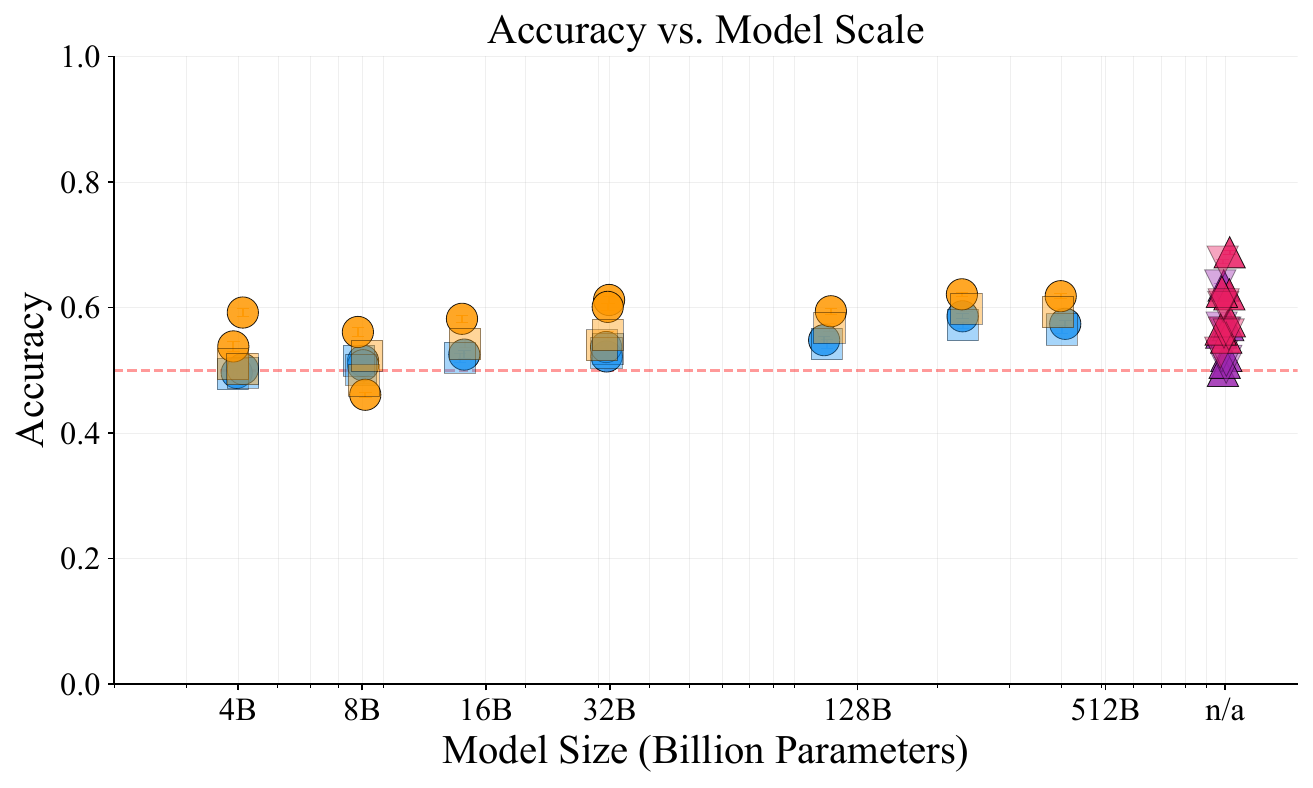}
    \end{subfigure}\hfill
    \begin{subfigure}[b]{0.5\linewidth}
    \includegraphics[width=\linewidth]{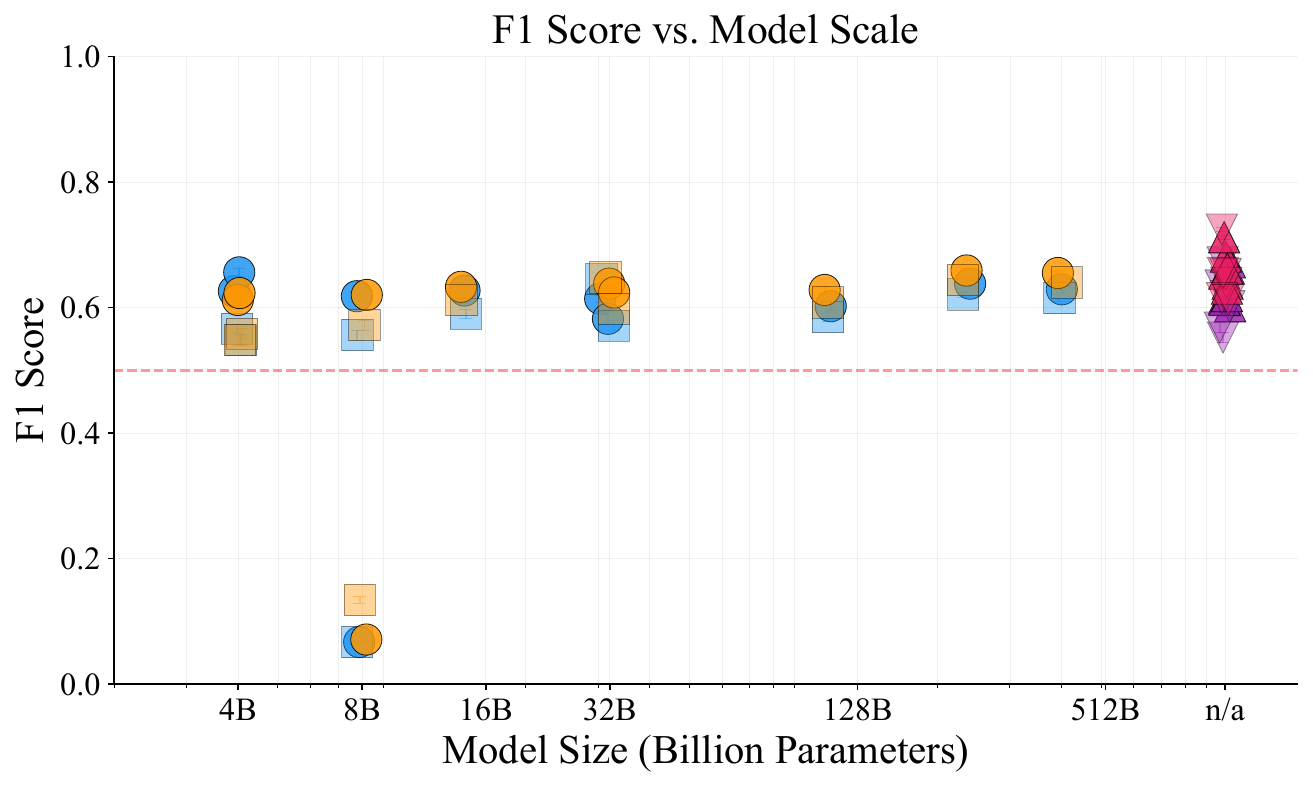}
    \end{subfigure}
    \caption{\textbf{AutoQA performance vs.\ model scale.} Accuracy (left) and F1 (right) on \datasetNameReal{} and \datasetNameSynthetic{} as a function of model size. Even the largest models remain below 70\% accuracy, indicating that visual consistency assessment is far from solved by current VLMs.
    }
    \label{fig:vllmperformance}
\end{figure}
}
\newcommand{\vllmfigureperclass}{
\begin{figure}[t!]
    \centering
   
    \begin{subfigure}[b]{0.5725\linewidth}
    \includegraphics[width=\linewidth]{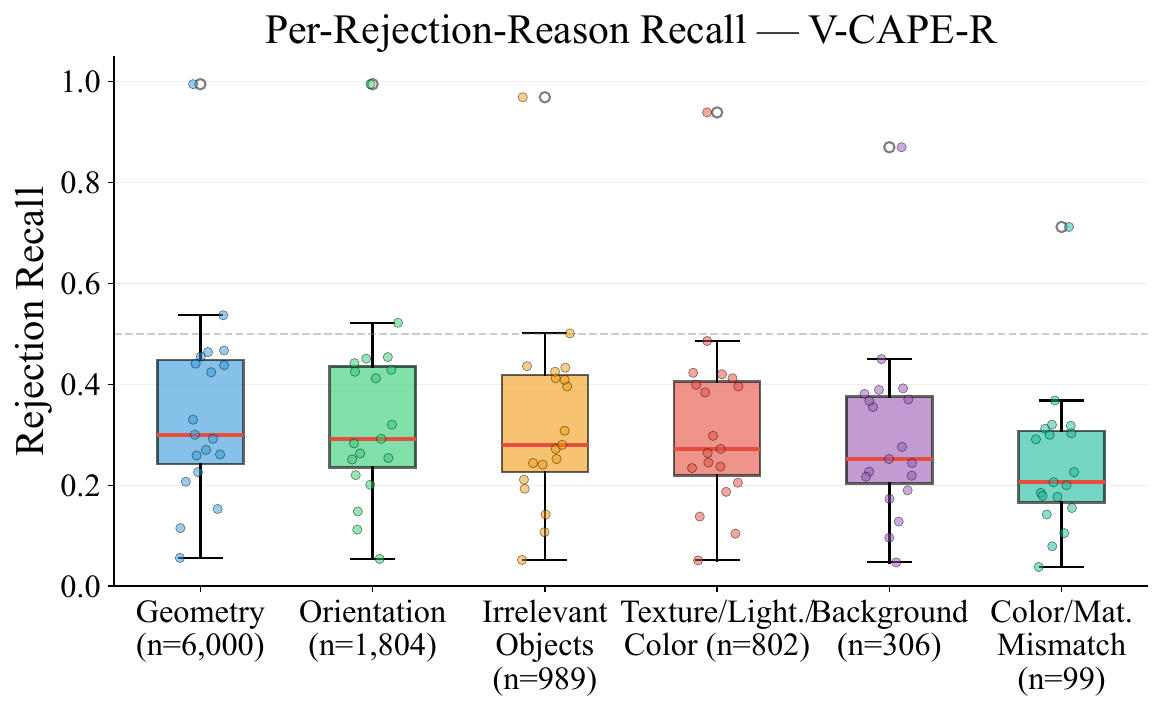}
    \end{subfigure}\hfill
    \begin{subfigure}[b]{0.4275\linewidth}
    \includegraphics[width=\linewidth]{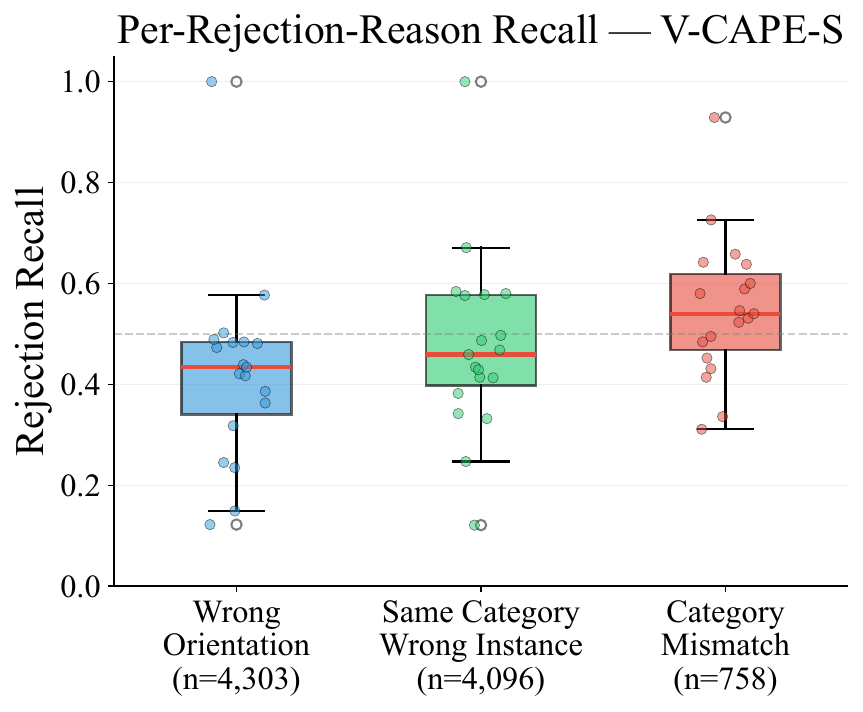}
    \end{subfigure}
    \caption{\textbf{AutoQA performance per-rejection reason.} Distribution of per-category rejection recall across all evaluated VLMs on \datasetNameReal{} (left) and \datasetNameSynthetic{} (right). Boxes: median and IQR; dots: individual models. Recall drops sharply from coarse violations (geometry or ca) to fine-grained mismatches (texture, lighting, background, colour/material), revealing a clear difficulty gradient that none of the evaluated models closes.
    }
    \label{fig:vllmperclassperformance}
\end{figure}
}

\newcommand{\vcapeexamples}{
\begin{figure}[t!]
    \centering
    
    \begin{subfigure}[t]{\linewidth}
        \centering
        \includegraphics[width=\linewidth]{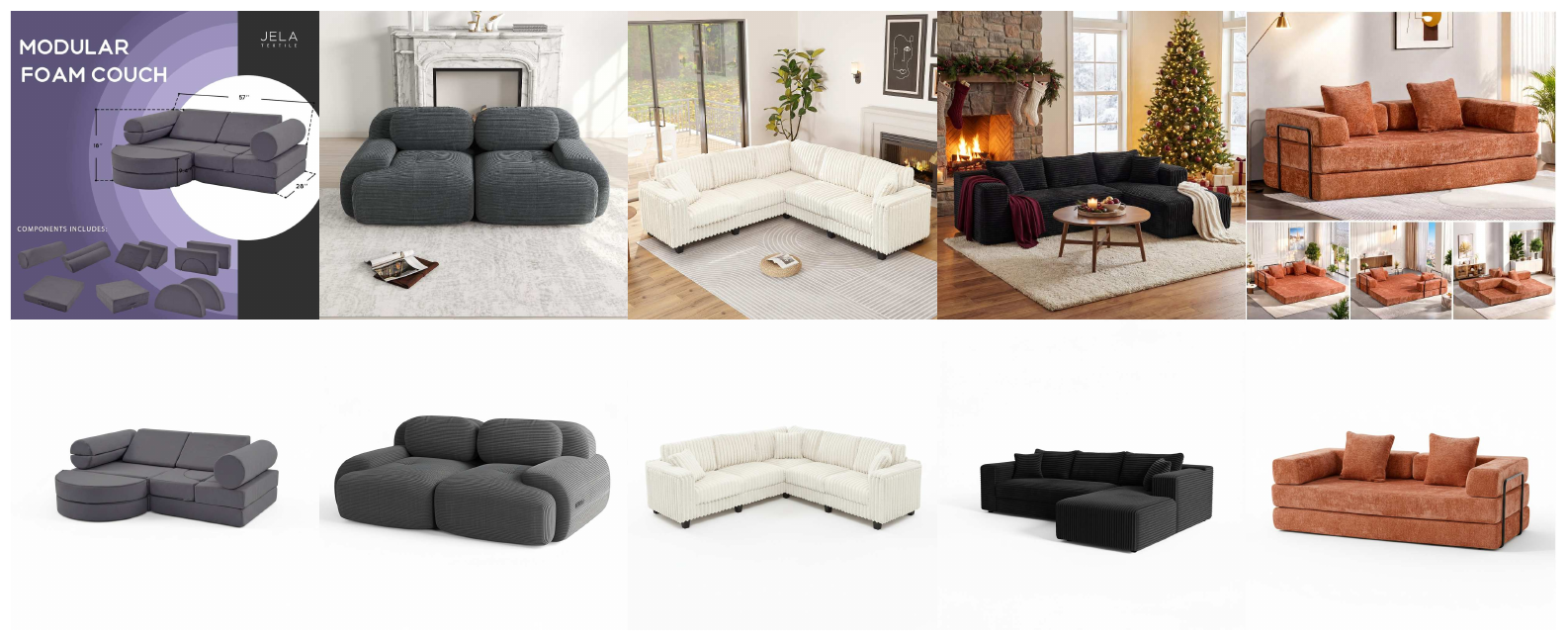}
        \caption{
        \textbf{\datasetNameReal.} Real-world generative outputs illustrating practical editing scenarios and associated visual consistency challenges.
        }
        \label{fig:vcape_r_pos_examples}
    \end{subfigure}
    \begin{subfigure}[t]{\linewidth}
        \centering
        \includegraphics[width=\linewidth]{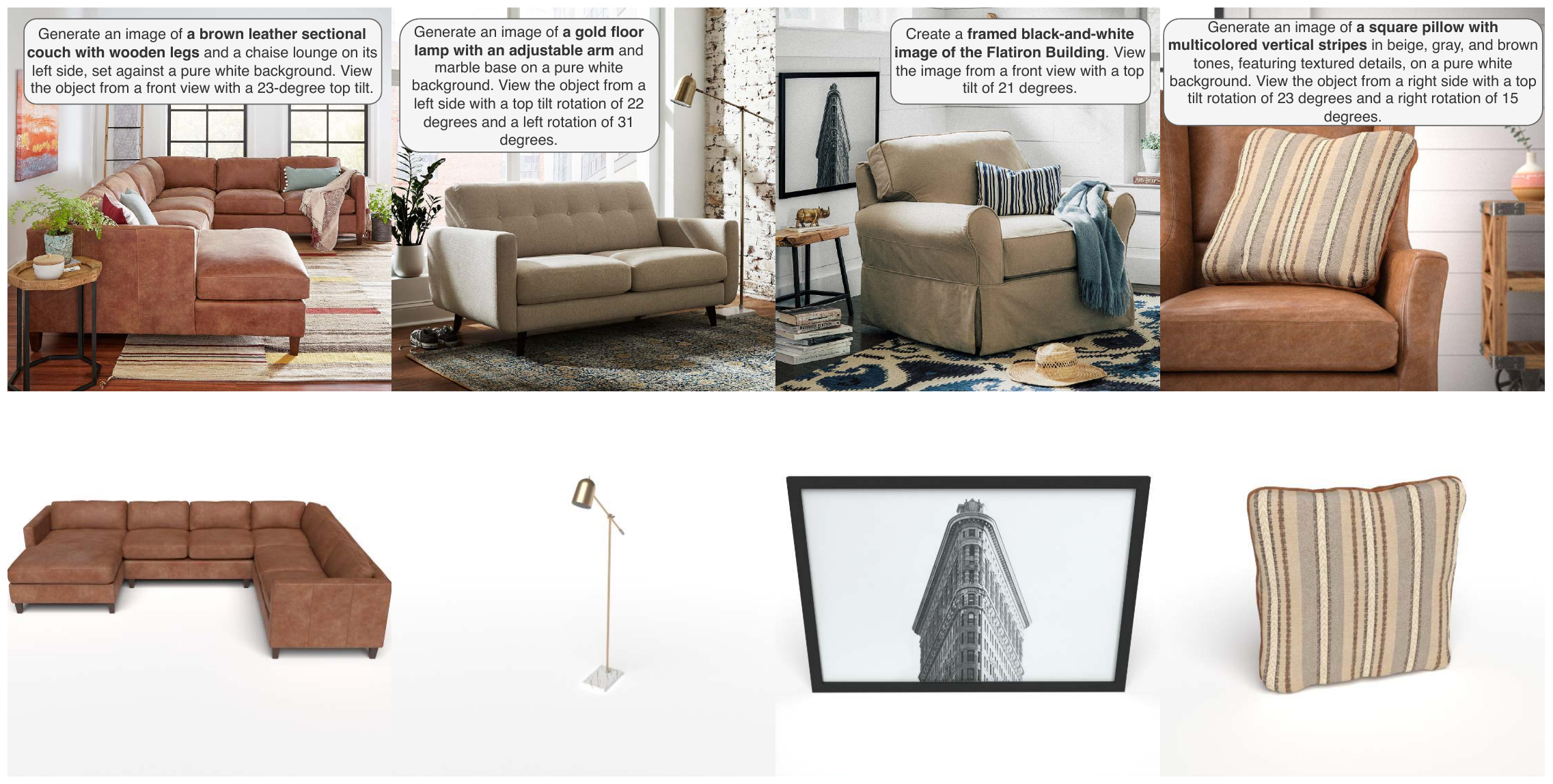}
        \caption{
        \textbf{\datasetNameSynthetic.} Each triplet consists of a source image (top), prompt, and target image (bottom), enabling controlled evaluation of geometry, pose, and fine-grained detail preservation.
        }
        \label{fig:vcape_s_pos_examples}
    \end{subfigure}
    \caption{
    \textbf{\datasetName overview.} Examples from the two complementary subsets: 
    (a) real-world generative outputs reflecting practical artifacts, and (b) synthetic data with controlled transformations and precise supervision. Together, they enable evaluation of fine-grained visual consistency under both realistic and controlled conditions.
    }
    \label{fig:vcape_combined}
\end{figure}
}

\newcommand{\plotvcapesstatssupp}{
    \pgfplotstableread[col sep=tab]{figures/vcape-s_pt_counts.dat}\datatable
    \begin{figure}[t!]
        \begin{tikzpicture}
        \centering
            \begin{axis}[
        ybar,
        width=\linewidth,
        height=5.7cm,
        ylabel={Number of Samples},
        ylabel style={font=\normalsize},
        ymin=0,
        scaled y ticks=base 10:-3,
        ymax=3200,
        ymajorgrids=true,
        ytick style={draw=none},
        xtick=data,
        xticklabels from table={\datatable}{label},
        x tick label style={
            rotate=45,
            anchor=east,
            font=\small,
            align=right,
            text width=2.5cm,
        },
        nodes near coords,
        nodes near coords style={
            fill=white,
            font=\tiny,
            /pgf/number format/1000 sep={},
        },
        every axis plot/.append style={
            fill=blue!60,
            draw=blue!80!black,
        },
        enlarge x limits={abs=0.4cm},
        clip=false,
        xtick pos=left,
    ]
    
    \addplot table [x expr=\coordindex, y=count] {\datatable};
    
    \end{axis}
    \end{tikzpicture}
    \vspace{-30pt}
    \caption{\textbf{\datasetNameSynthetic} product type distribution.}
    \end{figure}
}

\newcommand{\plotrejectedvcapessupp}{
    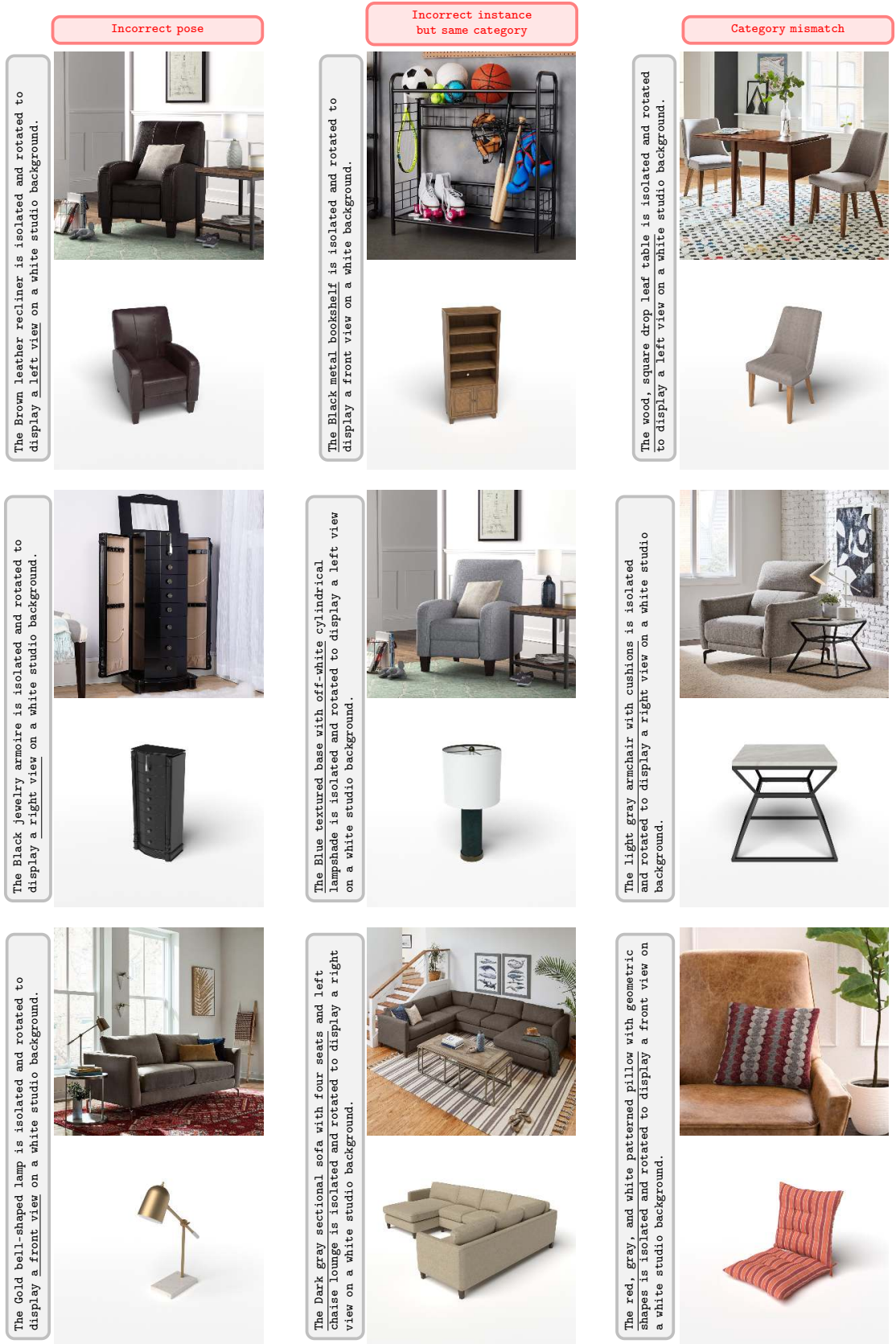
\begin{figure}[t!]
    \centering
        \begin{tikzpicture}[remember picture]
          \node[inner sep=0pt] (fig) at (0,0) {%
            \includegraphics[width=\textwidth]{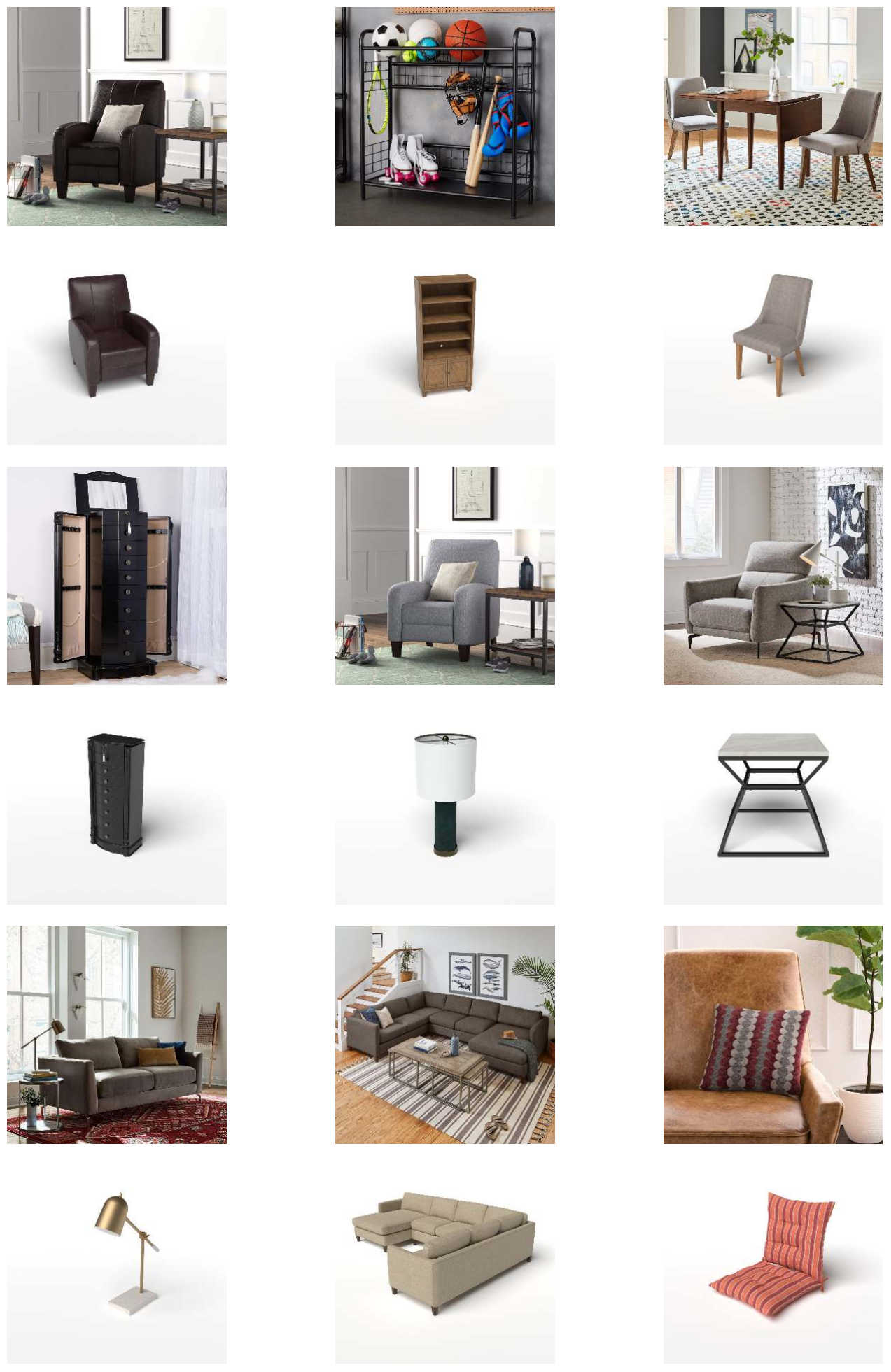}%
          };

          \node[
            anchor=center,
            rotate=90,
            inner sep=0pt,
          ] at (-7.3,7.25) {%
            \begin{tcolorbox}[
              colback=gray!10,
              colframe=gray!50,
              fontupper=\tiny\ttfamily,
              width=6.7cm,
              boxsep=2pt,
              left=1pt, right=1pt, top=1pt, bottom=1pt,
            ]
              The Brown leather recliner is isolated and rotated to display \underline{a left view} on a white studio background.
            \end{tcolorbox}%
          };

          \node[
            anchor=center,
            rotate=90,
            inner sep=0pt,
          ] at (-7.3,0.) {%
            \begin{tcolorbox}[
              colback=gray!10,
              colframe=gray!50,
              fontupper=\tiny\ttfamily,
              width=6.7cm,
              boxsep=2pt,
              left=1pt, right=1pt, top=1pt, bottom=1pt,
            ]
              The Black jewelry armoire is isolated and rotated to display \underline{a right view} on a white studio background.
            \end{tcolorbox}%
          };

          \node[
            anchor=center,
            rotate=90,
            inner sep=0pt,
          ] at (-7.3,-7.2) {%
            \begin{tcolorbox}[
              colback=gray!10,
              colframe=gray!50,
              fontupper=\tiny\ttfamily,
              width=6.7cm,
              boxsep=2pt,
              left=1pt, right=1pt, top=1pt, bottom=1pt,
            ]
              The Gold bell-shaped lamp is isolated and rotated to display \underline{a front view} on a white studio background.
            \end{tcolorbox}%
          };

          \node[
            anchor=center,
            rotate=90,
            inner sep=0pt,
          ] at (-2.15,7.25) {%
            \begin{tcolorbox}[
              colback=gray!10,
              colframe=gray!50,
              fontupper=\tiny\ttfamily,
              width=6.7cm,
              boxsep=2pt,
              left=1pt, right=1pt, top=1pt, bottom=1pt,
            ]
              \underline{The Black metal bookshelf} is isolated and rotated to display a front view on a white background.
            \end{tcolorbox}%
          };

          \node[
            anchor=center,
            rotate=90,
            inner sep=0pt,
          ] at (-2.25,0.) {%
            \begin{tcolorbox}[
              colback=gray!10,
              colframe=gray!50,
              fontupper=\tiny\ttfamily,
              width=6.7cm,
              boxsep=2pt,
              left=1pt, right=1pt, top=1pt, bottom=1pt,
            ]
              \underline{The Blue textured base with off-white} cylindrical lampshade is isolated and rotated to display a left view on a white studio background.
            \end{tcolorbox}%
          };

          \node[
            anchor=center,
            rotate=90,
            inner sep=0pt,
          ] at (-2.25,-7.2) {%
            \begin{tcolorbox}[
              colback=gray!10,
              colframe=gray!50,
              fontupper=\tiny\ttfamily,
              width=6.7cm,
              boxsep=2pt,
              left=1pt, right=1pt, top=1pt, bottom=1pt,
            ]
              \underline{The Dark gray sectional sofa with four seats} and left chaise lounge is isolated and rotated to display a right view on a white studio background.
            \end{tcolorbox}%
          };

        \node[
            anchor=center,
            rotate=90,
            inner sep=0pt,
          ] at (3,7.25) {%
            \begin{tcolorbox}[
              colback=gray!10,
              colframe=gray!50,
              fontupper=\tiny\ttfamily,
              width=6.7cm,
              boxsep=2pt,
              left=1pt, right=1pt, top=1pt, bottom=1pt,
            ]
              \underline{The wood, square drop leaf table} is isolated and rotated to display a left view on a white studio background.
            \end{tcolorbox}%
          };

          \node[
            anchor=center,
            rotate=90,
            inner sep=0pt,
          ] at (2.85,0.) {%
            \begin{tcolorbox}[
              colback=gray!10,
              colframe=gray!50,
              fontupper=\tiny\ttfamily,
              width=6.7cm,
              boxsep=2pt,
              left=1pt, right=1pt, top=1pt, bottom=1pt,
            ]
              \underline{The light gray armchair with cushions} is isolated and rotated to display a right view on a white studio background.
            \end{tcolorbox}%
          };

          \node[
            anchor=center,
            rotate=90,
            inner sep=0pt,
          ] at (2.85,-7.2) {%
            \begin{tcolorbox}[
              colback=gray!10,
              colframe=gray!50,
              fontupper=\tiny\ttfamily,
              width=6.7cm,
              boxsep=2pt,
              left=1pt, right=1pt, top=1pt, bottom=1pt,
            ]
              \underline{The red, gray, and white patterned pillow} with geometric shapes is isolated and rotated to display a front view on a white studio background.
            \end{tcolorbox}%
          };

          \node[
            anchor=center,
            inner sep=0pt,
          ] at (-5.17,11.) {%
            \begin{tcolorbox}[
              colback=red!10,
              colframe=red!50,
              fontupper=\tiny\ttfamily,
              halign=center,
              width=3.5cm,
              boxsep=2pt,
              left=1pt, right=1pt, top=1pt, bottom=1pt,
            ]
              \textcolor{red}{Incorrect pose}
            \end{tcolorbox}%
          };

          \node[
            anchor=center,
            inner sep=0pt,
          ] at (-0.,11.1) {%
            \begin{tcolorbox}[
              colback=red!10,
              colframe=red!50,
              fontupper=\tiny\ttfamily,
              halign=center,
              width=3.5cm,
              boxsep=2pt,
              left=1pt, right=1pt, top=1pt, bottom=1pt,
            ]
              \textcolor{red}{Incorrect instance but same category}
            \end{tcolorbox}%
          };

          \node[
            anchor=center,
            inner sep=0pt,
          ] at (5.2,11.) {%
            \begin{tcolorbox}[
              colback=red!10,
              colframe=red!50,
              fontupper=\tiny\ttfamily,
              halign=center,
              width=3.5cm,
              boxsep=2pt,
              left=1pt, right=1pt, top=1pt, bottom=1pt,
            ]
              \textcolor{red}{Category mismatch}
            \end{tcolorbox}%
          };
               
        \end{tikzpicture}
    \caption{\textbf{\datasetNameSynthetic rejected pairs}. For each rejection category in our taxonomy, we show the source--target image pair alongside the corresponding editing prompt.}\label{s_fig:vcape-s-quali}
\end{figure}
}

\newcommand{\plotrejectedvcaperIsupp}{
    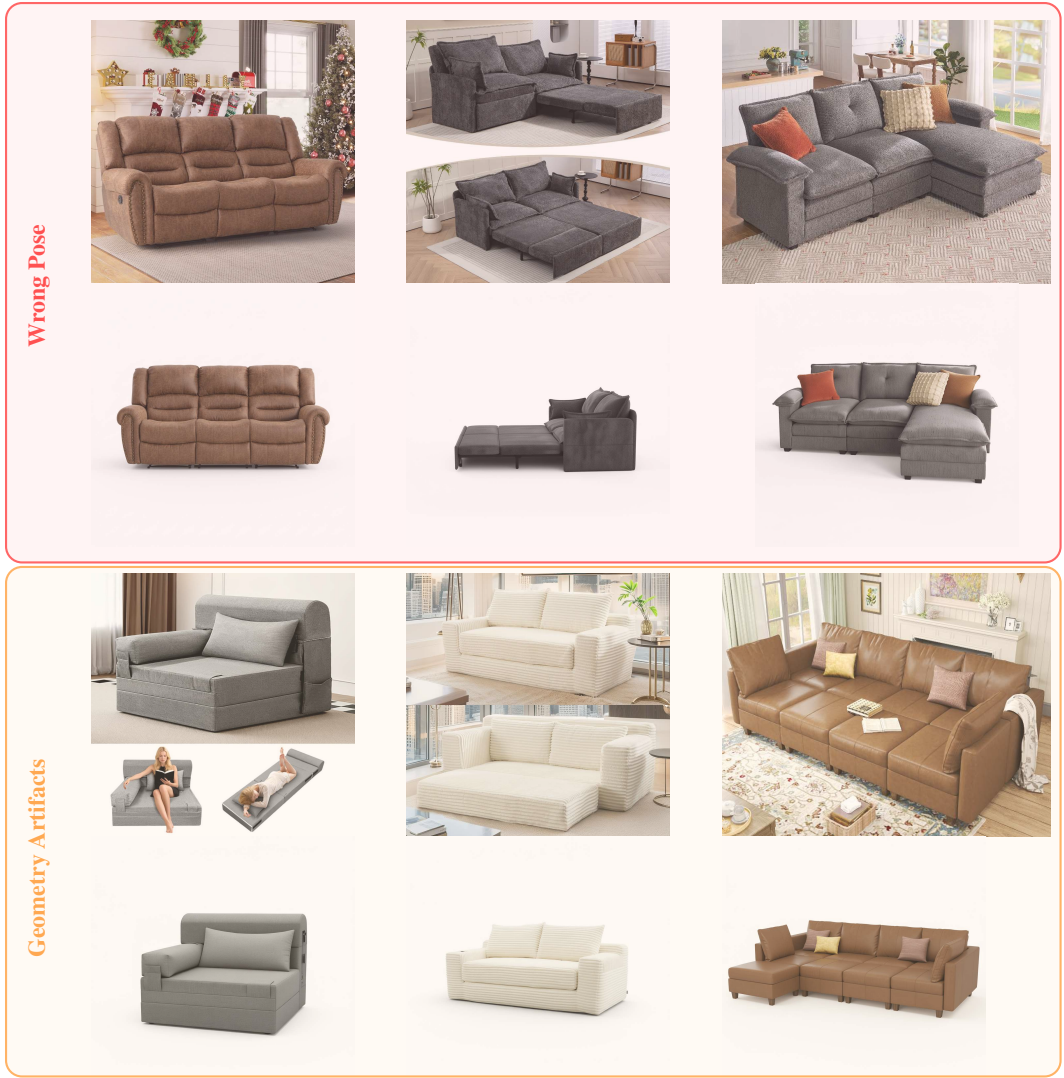
\begin{figure}[t!]
    \centering
        \begin{tikzpicture}[remember picture]
          \node[inner sep=0pt] (fig) at (0,0) {%
            \includegraphics[width=.926\textwidth]{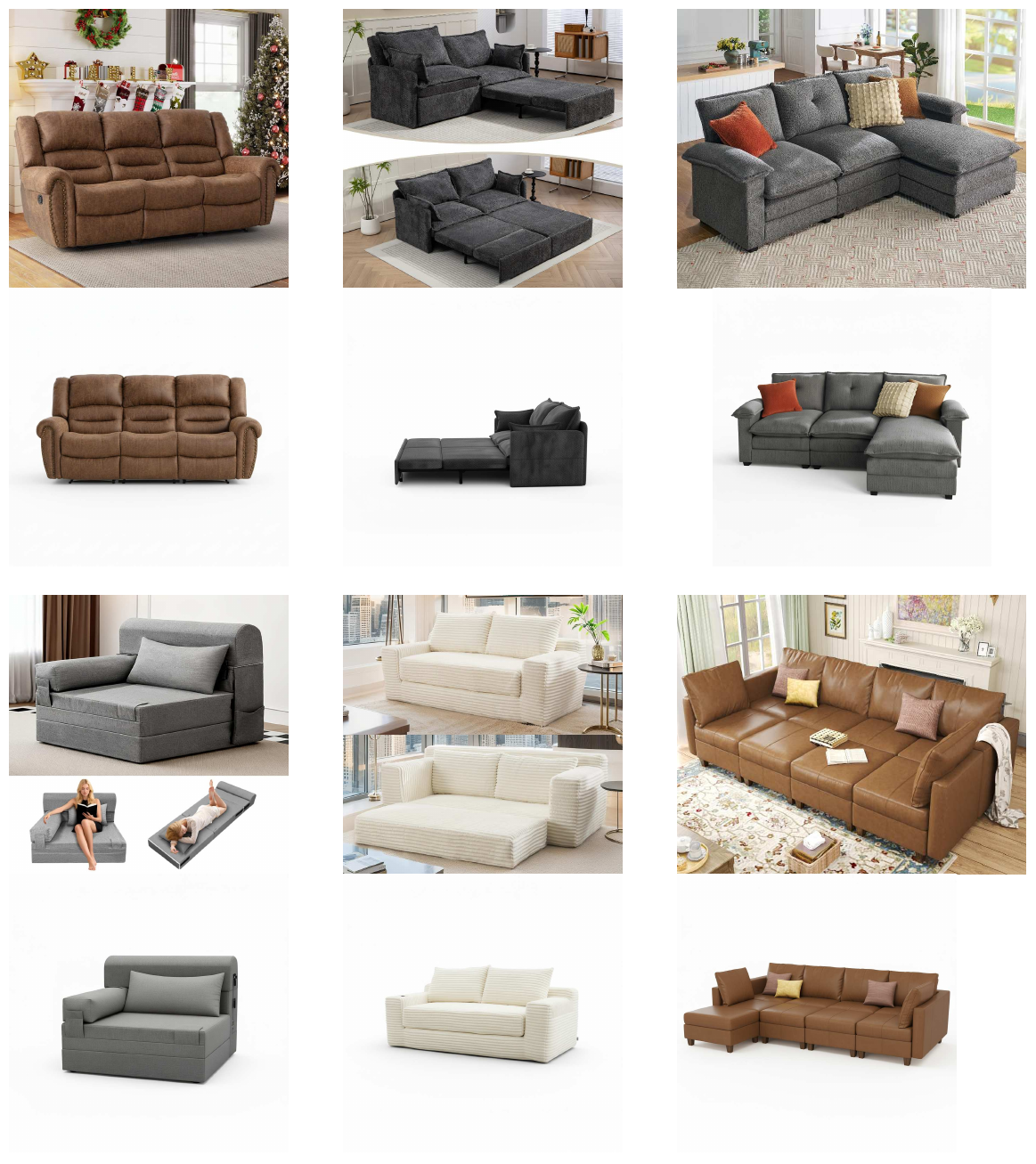}%
          };

          \node[
              rotate=90,
              anchor=south,
              font=\small\bfseries,
              text=red!80
            ]
            at ($(fig.north west)!0.25!(fig.south west) + (-0.3cm,0)$)
            {\textbf{Wrong Pose}};
          \draw[
              rounded corners=6pt,
              draw=red!60,
              fill=red!20,
              fill opacity=0.2,
              line width=0.8pt
            ]
            ($(fig.north west) + (-1cm,0.1cm)$) rectangle
            ($(fig.north east) + (0,-7.3cm)$);

        \node[
              rotate=90,
              anchor=south,
              font=\small\bfseries,
              text=orange!80
            ]
            at ($(fig.north west)!0.77!(fig.south west) + (-0.3cm,0)$)
            {\textbf{Geometry Artifacts}};
        \draw[
              rounded corners=6pt,
              draw=orange!60,
              fill=orange!20,
              fill opacity=0.2,
              line width=0.8pt
            ]
            ($(fig.north west) + (-1cm,-7.36cm)$) rectangle
            ($(fig.north east) + (0,-14.1cm)$);

        \end{tikzpicture}
    \caption{ \textbf{Visual overview of the \datasetNameReal rejection taxonomy}. We showcase representative failure cases, including \textcolor{red}{\textbf{Wrong Pose}} and  \textcolor{orange}{\textbf{Geometry Artifacts}}, highlighting the fine-grained error categories (\cf~\tabref{ssup:tab_taxonomy}) captured by our annotations.}\label{s_fig:vcape-r-quali-2}
\end{figure}
}

\newcommand{\plotrejectedvcaperIIsupp}{
    \begin{figure}[t!]
    \centering
        \begin{tikzpicture}[remember picture]
          \node[inner sep=0pt] (fig) at (0,0) {%
            \includegraphics[width=.926\textwidth]{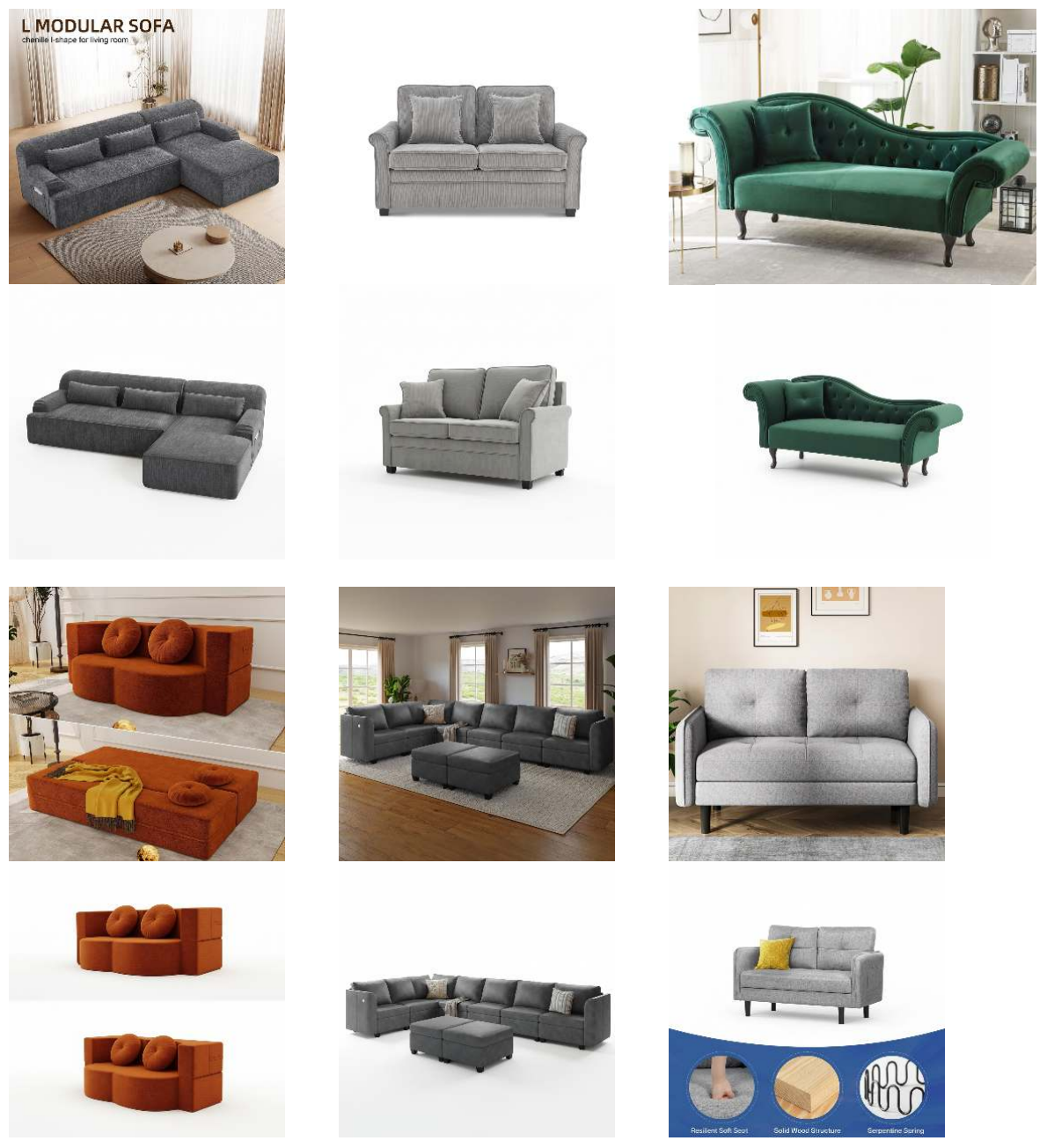}%
          };

          \node[
              rotate=90,
              anchor=south,
              font=\small\bfseries,
              text=brown!80
            ]
            at ($(fig.north west)!0.26!(fig.south west) + (-0.3cm,0)$)
            {\textbf{Texture / Lighting / Color Issues}};
          \draw[
              rounded corners=6pt,
              draw=brown!60,
              fill=brown!20,
              fill opacity=0.2,
              line width=0.8pt
            ]
            ($(fig.north west) + (-1cm,0.1cm)$) rectangle
            ($(fig.north east) + (0,-7.0cm)$);

        \node[
              rotate=90,
              anchor=south,
              font=\small\bfseries,
              text=amethyst!80
            ]
            at ($(fig.north west)!0.768!(fig.south west) + (-0.3cm,0)$)
            {\textbf{Background Issues}};
        \draw[
              rounded corners=6pt,
              draw=amethyst!60,
              fill=amethyst!20,
              fill opacity=0.2,
              line width=0.8pt
            ]
            ($(fig.north west) + (-1cm,-7.1cm)$) rectangle
            ($(fig.north east) + (0,-14.2cm)$);

        \end{tikzpicture}
    \caption{ \textbf{Visual overview of the \datasetNameReal rejection taxonomy}. We showcase representative failure cases, including \textcolor{brown}{\textbf{Texture / Lighting / Color Issues}} and  \textcolor{amethyst}{\textbf{Background Issues}}, highlighting the fine-grained error categories (\cf~\tabref{ssup:tab_taxonomy}) captured by our annotations.}\label{s_fig:vcape-r-quali-3}
\end{figure}
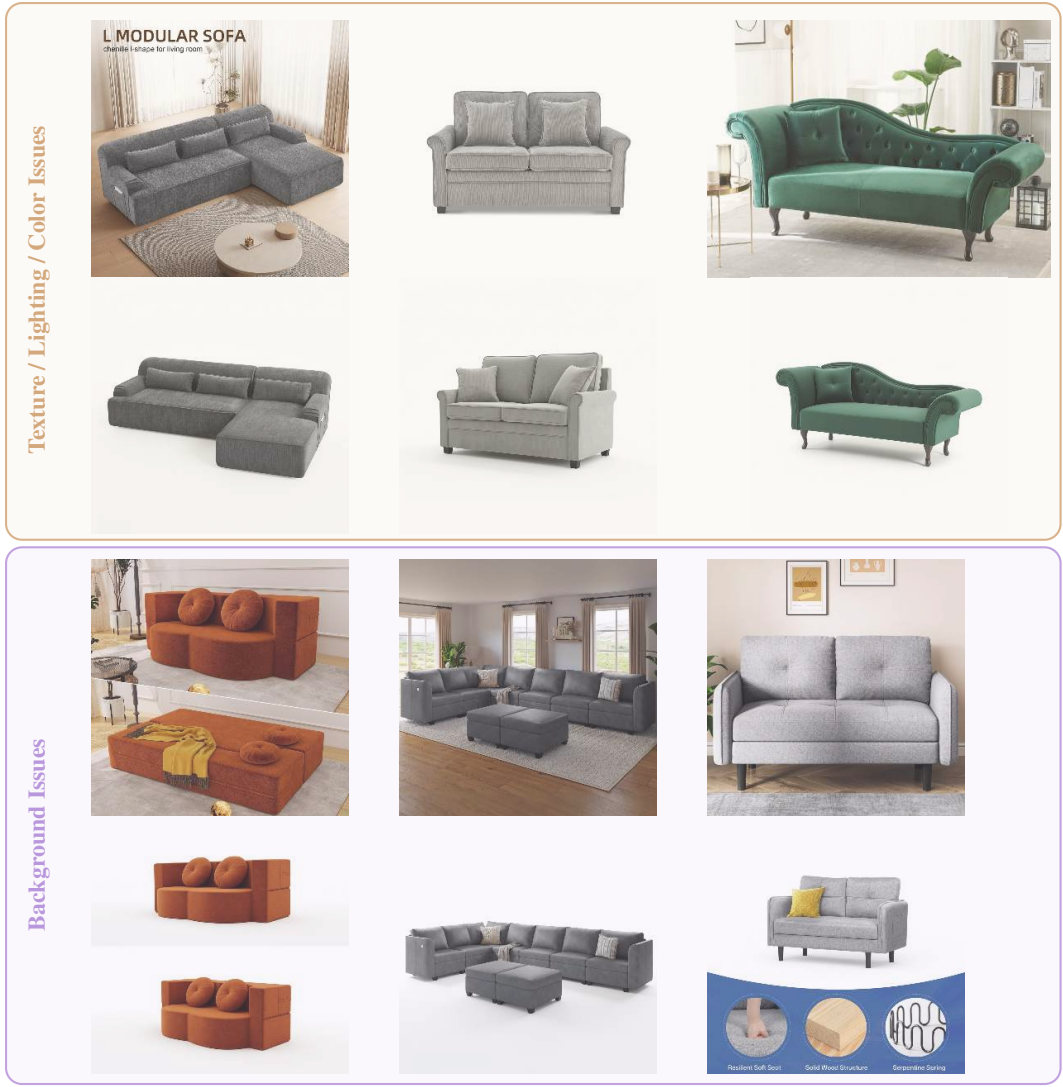
}

\newcommand{\plotrejectedvcaperIIIsupp}{
    \begin{figure}[t!]
    \centering
        \begin{tikzpicture}[remember picture]
          \node[inner sep=0pt] (fig) at (0,0) {%
            \includegraphics[width=.926\textwidth]{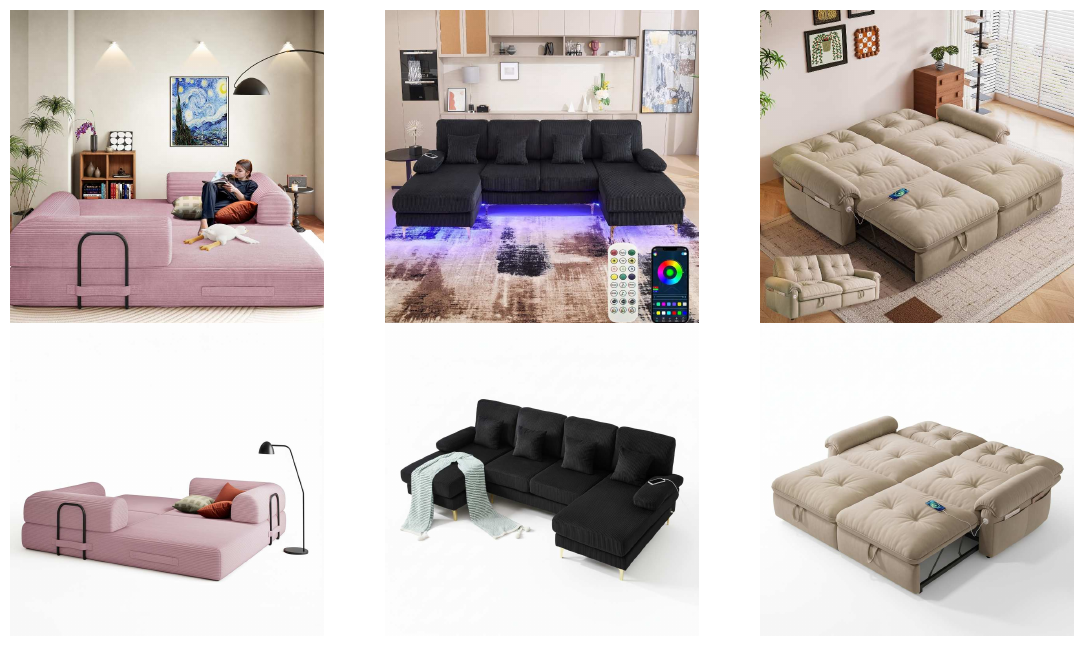}%
          };

          \node[
              rotate=90,
              anchor=south,
              font=\small\bfseries,
              text=blue!80
            ]
            at ($(fig.north west)!0.5!(fig.south west) + (-0.3cm,0)$)
            {\textbf{Irrelevant Objects}};
          \draw[
              rounded corners=6pt,
              draw=blue!60,
              fill=blue!20,
              fill opacity=0.2,
              line width=0.8pt
            ]
            ($(fig.north west) + (-1cm,0.1cm)$) rectangle
            ($(fig.north east) + (0,-7.9cm)$);

        \end{tikzpicture}
    \caption{\textbf{\datasetNameReal rejected pairs} for the \textcolor{blue}{\textbf{Irrelevant Objects}} category (\cf~\tabref{ssup:tab_taxonomy}). Unlike prior datasets, our annotations provide explicit rejection reasons for each sample.}\label{s_fig:vcape-r-irrelevant-examples}
\end{figure}
}

\newcommand{\plotsourceandgenerationdetailed}{
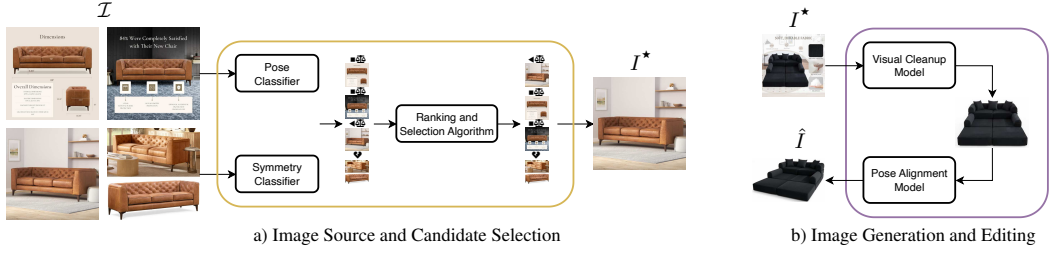
\begin{figure}[t!]
    \centering
    \begin{tikzpicture}[
        POI/.style={circle, fill=red, inner sep=1pt},
        POI label/.append style={font=\footnotesize,inner sep=0.1em},
        main label/.append style={font=\scriptsize,inner sep=0.1em},
        inner label/.append style={font=\fontsize{3.0}{5}\selectfont,inner sep=0.1em},
        ]
        
        \node[inner sep=0pt]
         {\includegraphics[width=\linewidth,trim=0 0 0 0 ,clip]{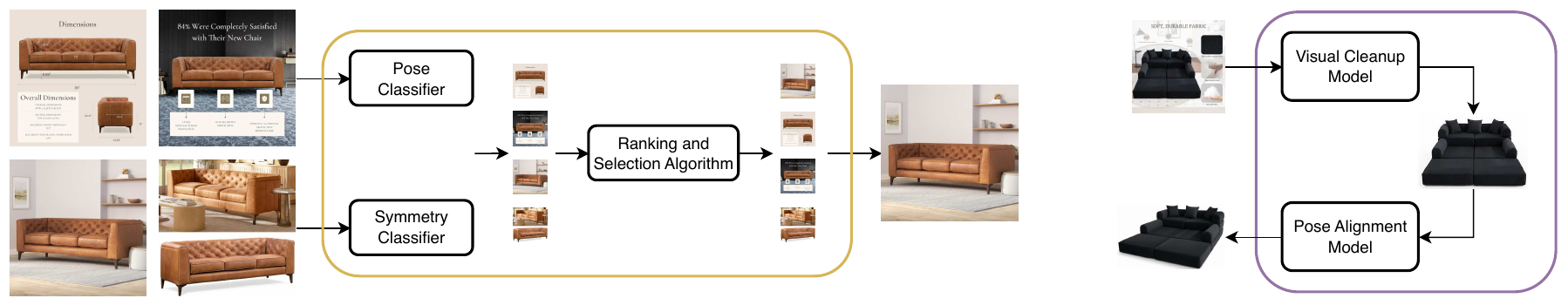}};

         \path (-5.6, 1.5) node[POI label] {$\mathcal{I}$};
         \path (1.5, 0.83) node[POI label] {$I^\star$};
         \path (3.6, 1.4) node[POI label] {$I^\star$};
         \path (3.6, -0.2) node[POI label] {$\hat{I}$};

         \path (-2.25, 0.85) node[inner label] {$\medblacksquare$\faBalanceScale};
         \path (-2.25, 0.43) node[inner label] {$\medblacksquare$\faBalanceScale};
         \path (-2.26, 0.01) node[inner label] {$\medblacktriangleleft$\faBalanceScale};
         \path (-2.25, -0.42) node[inner label] {\faHeartBroken};

         \path (0.11, 0.85) node[inner label] {$\medblacktriangleleft$\faBalanceScale};
         \path (0.11, 0.42) node[inner label] {$\medblacksquare$\faBalanceScale};
         \path (0.11, 0.01) node[inner label] {$\medblacksquare$\faBalanceScale};
         \path (0.11, -0.42) node[inner label] {\faHeartBroken};

         \path (-1.6, -1.5) node[main label] {a) Image Source and Candidate Selection};
         \path (5.1, -1.5) node[main label] {b) Image Generation and Editing};

    \end{tikzpicture}
    \caption{\textbf{Image candidate selection and generation modules.} a) Given a set of catalog images $\mathcal{I}$ for a given product, each image is independently processed by pose and symmetry classifiers to infer object orientation and symmetry attributes (\cf~\secref{ssec:image_sourcing}). These attributes, together with predefined visual consistency constraints, are used by a ranking and selection algorithm to deterministically select a single source image $I^\star$ for downstream processing. b) Image generation and editing: Starting from $I^\star$, background clutter is first removed using Gemini 2.5 Flash~\cite{nano_banana} while preserving the main object. The QEM model~\cite{qwen_edit_multiple_angles} is then applied to perform controlled geometric edits, aligning the object pose with the target visual consistency requirements.
}\label{fig:image_selection_and_generation}
\end{figure}
}

\newcommand{\plotrankingacceptancerate}{
\begin{wrapfigure}{r}{0.4\linewidth}
    \centering
    \vspace{-0.5\baselineskip}
    \begin{tikzpicture}
\begin{axis}[
    xbar,
    width=10cm,
    height=5cm,
    xlabel={Acceptance Rate (\%)},
    symbolic y coords={$\medblacktriangleright$\faBalanceScaleLeft, $\medblacksquare$, $\medblackdiamond$, $\medblacktriangleright$\faBalanceScale, $\medblacktriangleleft$},
    ytick=data,
    xmin=50, xmax=85,
    nodes near coords={\pgfmathprintnumber\pgfplotspointmeta},
    nodes near coords align={horizontal},
    bar width=10pt,
]
\addplot[fill=blue!60!white, draw=blue!70!black] coordinates {
    (54.88,$\medblacktriangleright$\faBalanceScaleLeft)
    (57.94,$\medblacksquare$)
    (76.67,$\medblackdiamond$)
    (79.56,$\medblacktriangleright$\faBalanceScale)
    (80.44,$\medblacktriangleleft$)
};
\end{axis}
\end{tikzpicture}
    \vspace{-1.5\baselineskip}
    \caption{\textbf{Acceptance rate for image selection}. Left-facing $(\medblacktriangleleft)$ and right-symmetric $(\medblacktriangleright$\faBalanceScale$)$ cases achieve the highest acceptance rates ($>$79\%).}
    \label{fig:ranking_acceptance_rates}
    \vspace{-10pt}
\end{wrapfigure}
}

\newcommand{\annotationtemplate}{
\begin{figure}[t!]
    \centering
    \includegraphics[width=\linewidth]{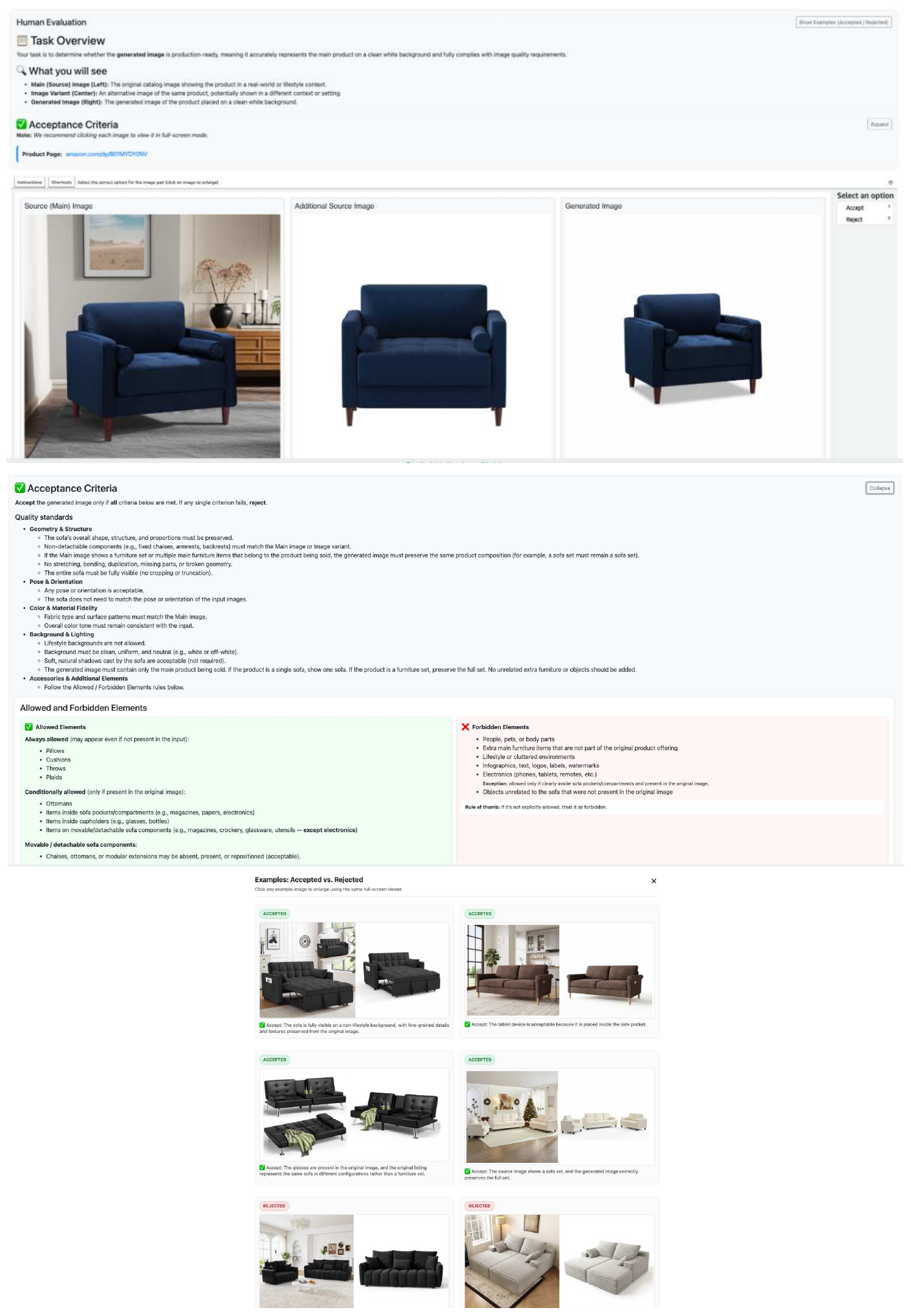}
    \caption{
        \textbf{Annotation template.} We employ this template to collect labels from three annotators per image triplet (source, additional view, and target), enabling the assessment of generated image quality according to the proposed label taxonomy (\tabref{ssup:tab_taxonomy}).
    }
    \label{fig:annotation_template}
\end{figure}
}

\newcommand{\finetuningcurves}{
\begin{figure}[b!]
    \centering
    \includegraphics[width=\linewidth]{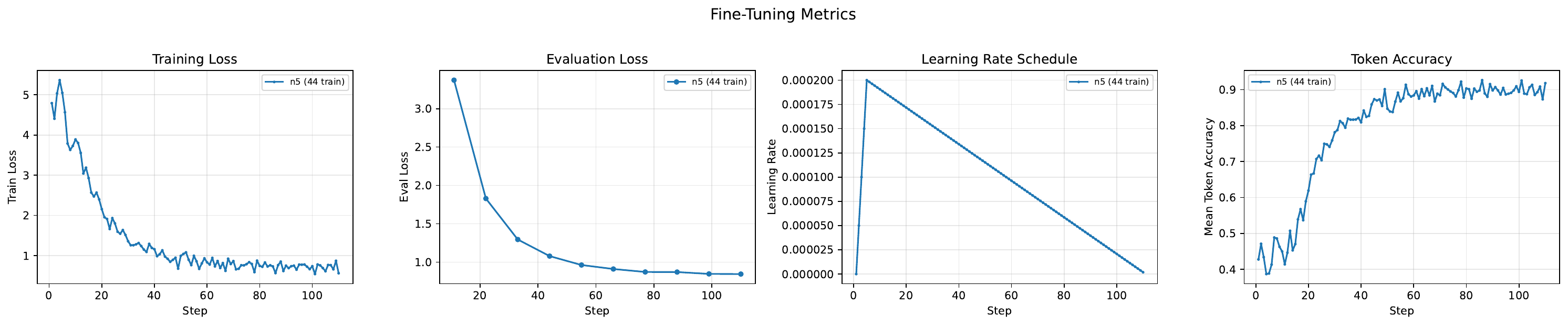}
    \caption{
        \textbf{\datasetName fine-tuning: training dynamics} %
    }
    \label{fig:finetune_dynamics}
\end{figure}
}

\newcommand{\zeroshotorientation}{
  \begin{table*}[t]
    \centering
    \small
    \setlength{\tabcolsep}{2.5pt}
    \begin{tabular}{llcccccccc  c}
    \toprule
    & & \multicolumn{4}{c}{\datasetNameSynthetic (3-class)} & & \multicolumn{3}{c}{\datasetNameReal (2-class)} \\
    \cmidrule(lr){3-6} \cmidrule(lr){8-10}
    & & & \multicolumn{3}{c}{F1} & & & \multicolumn{2}{c}{F1} \\
    \cmidrule(lr){4-6} \cmidrule(lr){9-10}
    Model & Proto. & Acc. & Front & Left & Right & & Acc. & Front & Left \\
    \midrule
    \multicolumn{10}{l}{\textit{Contrastive vision-language models}} \\
    CLIP~\cite{radford2021clip}         & image/text & \mstd{.335}{.004} & \mstd{.392}{.006} & \mstd{.354}{.006} & \mstd{.204}{.006} & & \mstd{.611}{.005} & \mstd{.158}{.007} & \mstd{.747}{.004} \\
    EVA-02-CLIP~\cite{EVA-CLIP}  & image/text & \mstd{.341}{.004} & \mstd{.481}{.005} & \mstd{.166}{.006} & \mstd{.163}{.006} & & \mstd{.548}{.005} & \mstd{.112}{.006} & \mstd{.697}{.004} \\
    SigLIP2~\cite{tschannen2025siglip2}      & image/text & \mstd{.471}{.004} & \mstd{.587}{.005} & \mstd{.536}{.005} & \mstd{.003}{.001} & & \mstd{.716}{.004} & \mstd{.130}{.008} & \mstd{.830}{.003} \\
    \midrule
    \multicolumn{10}{l}{\textit{Self-supervised vision encoders ($N{=}20$ shots/class)}} \\
    DINOv2~ViT-S~\cite{oquab2024dinov2}       & image & \mstd{.568}{.004} & \mstd{.558}{.006} & \mstd{.584}{.006} & \textbf{\mstd{.561}{.006}} & & \mstd{.758}{.004} & \mstd{.477}{.009} & \mstd{.842}{.003} \\
    DINOv2~ViT-B~\cite{oquab2024dinov2}       & image & \mstd{.449}{.004} & \mstd{.488}{.006} & \mstd{.436}{.006} & \mstd{.419}{.006} & & \mstd{.748}{.004} & \mstd{.455}{.009} & \mstd{.836}{.003} \\
    DINOv2~ViT-L~\cite{oquab2024dinov2}       & image & \mstd{.391}{.004} & \mstd{.418}{.006} & \mstd{.417}{.006} & \mstd{.334}{.006} & & \mstd{.752}{.004} & \mstd{.446}{.009} & \mstd{.841}{.003} \\
    EUPE-B~\cite{zhu2026eupe}         & image & \mstd{.332}{.004} & \mstd{.446}{.005} & \mstd{.231}{.006} & \mstd{.195}{.006} & & \mstd{.757}{.004} & \mstd{.449}{.009} & \mstd{.844}{.003} \\
    \midrule
    \multicolumn{10}{l}{\textit{Orientation estimation models (calibrated azimuth boundaries)}} \\
    Orient\-Anything~V1~\cite{orient_anything}      & azimuth & \mstd{.586}{.004} & \mstd{.578}{.006} & \mstd{.632}{.005} & \mstd{.530}{.006} & & \mstd{.855}{.004} & \mstd{.514}{.011} & \mstd{.917}{.002} \\
    Orient\-Anything V2~\cite{wangorientv2}      & azimuth & \textbf{\mstd{.590}{.004}} & \textbf{\mstd{.606}{.005}} & \textbf{\mstd{.683}{.005}} & \mstd{.415}{.007} & & \textbf{\mstd{.911}{.003}} & \textbf{\mstd{.622}{.012}} & \textbf{\mstd{.951}{.002}} \\
    \bottomrule
    \end{tabular}
    \caption{\textbf{Zero-shot orientation classification results}. We report orientation classification accuracy and per-class F1-score (bootstrap mean $\pm$ std, $B{=}1{,}000$) for \datasetNameSynthetic and \datasetNameReal.} %
    \label{tab:orient}
    \end{table*}
    
}
 \newcommand{\vllmcomparison}{
  \begin{table*}[t]
  \centering
  \small
  \setlength{\tabcolsep}{2.2pt}
  \begin{tabular}{ll cccc cccc}
  \toprule
  & & \multicolumn{4}{c}{\datasetNameReal{}} & \multicolumn{4}{c}{\datasetNameSynthetic{}} \\
  \cmidrule(lr){3-6} \cmidrule(lr){7-10}
  & & \multicolumn{2}{c}{Baseline} & \multicolumn{2}{c}{+ Auto-prompt~\cite{levi2024ipc}} & \multicolumn{2}{c}{Baseline} & \multicolumn{2}{c}{+ Auto-prompt~\cite{levi2024ipc}} \\
  \cmidrule(lr){3-4} \cmidrule(lr){5-6} \cmidrule(lr){7-8} \cmidrule(lr){9-10}
  Model & Params & Acc. & F1 & Acc. & F1 & Acc. & F1 & Acc. & F1 \\
  \midrule
  \multicolumn{10}{l}{\textit{Open-weight models}} \\
  Gemma~3            & 4B    & \mstd{.496}{.007} & \mstd{.626}{.008} & \mstd{.538}{.008} & \mstd{.612}{.009} & \mstd{.494}{.006} & \mstd{.566}{.007} & \mstd{.510}{.007} & \mstd{.558}{.008} \\
  Gemma~4            & 4B    & \mstd{.502}{.006} & \mstd{.656}{.007} & \mstd{.592}{.006} & \mstd{.623}{.007} & \mstd{.496}{.008} & \mstd{.548}{.009} & \mstd{.502}{.007} & \mstd{.548}{.008} \\
  Gemma~4            & 31B   & \mstd{.522}{.005} & \mstd{.614}{.006} & \mstd{.612}{.005} & \mstd{.638}{.005} & \mstd{.534}{.006} & \mstd{.645}{.006} & \mstd{.540}{.005} & \mstd{.648}{.006} \\
  Qwen~3             & 8B    & \mstd{.515}{.003} & \mstd{.067}{.003} & \mstd{.461}{.004} & \mstd{.071}{.004} & \mstd{.515}{.003} & \mstd{.067}{.002} & \mstd{.484}{.005} & \mstd{.134}{.005} \\
  Qwen~3$_{\text{LoRA-n1}}$  & 8B  & \mstd{.521}{-}    & \mstd{.601}{-}    & \mstd{x}{-}       & \mstd{x}{-}       & \mstd{.532}{-}    & \mstd{.624}{-}    & \mstd{x}{-}       & \mstd{x}{-} \\
  Qwen~3$_{\text{LoRA-n5}}$  & 8B  & \mstd{.529}{-}    & \mstd{.628}{-}    & \mstd{x}{-}       & \mstd{x}{-}       & \mstd{.551}{-}    & \mstd{.683}{-}    & \mstd{x}{-}       & \mstd{x}{-} \\
  Qwen~3$_{\text{LoRA-n10}}$ & 8B  & \mstd{.527}{-}    & \mstd{.617}{-}    & \mstd{x}{-}       & \mstd{x}{-}       & \mstd{.543}{-}    & \mstd{.681}{-}    & \mstd{x}{-}       & \mstd{x}{-} \\
  Qwen~3             & 32B   & \mstd{.537}{.006} & \mstd{.582}{.007} & \mstd{.601}{.005} & \mstd{.624}{.006} & \mstd{.528}{.007} & \mstd{.571}{.008} & \mstd{.556}{.006} & \mstd{.598}{.006} \\
  Qwen~3             & 235B  & \mstd{.586}{.004} & \mstd{.638}{.004} & \mstd{.621}{.003} & \mstd{.659}{.004} & \mstd{.572}{.005} & \mstd{.621}{.005} & \mstd{.598}{.004} & \mstd{.644}{.004} \\
  Llama~4 Scout      & 109B  & \mstd{.548}{.006} & \mstd{.602}{.007} & \mstd{.594}{.005} & \mstd{.628}{.005} & \mstd{.542}{.007} & \mstd{.585}{.007} & \mstd{.568}{.006} & \mstd{.608}{.006} \\
  Llama~4 Maverick   & 400B  & \mstd{.574}{.005} & \mstd{.629}{.005} & \mstd{.618}{.004} & \mstd{.655}{.004} & \mstd{.565}{.006} & \mstd{.615}{.006} & \mstd{.593}{.005} & \mstd{.640}{.005} \\
  Mistral~3          & 8B    & \mstd{.508}{.007} & \mstd{.618}{.008} & \mstd{.561}{.007} & \mstd{.620}{.008} & \mstd{.501}{.008} & \mstd{.556}{.008} & \mstd{.523}{.007} & \mstd{.572}{.008} \\
  Mistral~3          & 14B   & \mstd{.525}{.006} & \mstd{.627}{.007} & \mstd{.582}{.006} & \mstd{.633}{.006} & \mstd{.520}{.007} & \mstd{.590}{.007} & \mstd{.542}{.006} & \mstd{.612}{.006} \\
  \midrule
  \multicolumn{10}{l}{\textit{Proprietary models}} \\
  Claude Sonnet~3.7  & n/a   & \mstd{.523}{.005} & \mstd{.662}{.005} & \mstd{.578}{.006} & \mstd{.681}{.006} & \mstd{.522}{.006} & \mstd{.635}{.007} & \mstd{.528}{.007} & \mstd{.639}{.007} \\
  Claude Sonnet~4.5  & n/a   & \mstd{.560}{.004} & \mstd{.602}{.005} & \mstd{.621}{.004} & \mstd{.654}{.004} & \mstd{.560}{.005} & \mstd{.602}{.006} & \mstd{.601}{.004} & \mstd{.653}{.005} \\
  Claude Haiku~4.5   & n/a   & \mstd{.499}{.007} & \mstd{.614}{.008} & \mstd{.552}{.006} & \mstd{.638}{.007} & \mstd{.504}{.007} & \mstd{.552}{.008} & \mstd{.560}{.006} & \mstd{.615}{.006} \\
  Claude Opus~4.5    & n/a   & \textbf{\mstd{.635}{.003}} & \textbf{\mstd{.672}{.003}} & \textbf{\mstd{.688}{.003}} & \textbf{\mstd{.712}{.003}} & \textbf{\mstd{.635}{.004}} & \textbf{\mstd{.672}{.004}} &
  \textbf{\mstd{.673}{.003}} & \textbf{\mstd{.724}{.003}} \\
  Nova Lite~2        & n/a   & \mstd{.512}{.007} & \mstd{.612}{.008} & \mstd{.565}{.006} & \mstd{.630}{.007} & \mstd{.515}{.007} & \mstd{.568}{.008} & \mstd{.557}{.006} & \mstd{.612}{.006} \\
  Nova Pro           & n/a   & \mstd{.572}{.005} & \mstd{.618}{.005} & \mstd{.625}{.004} & \mstd{.662}{.004} & \mstd{.568}{.005} & \mstd{.618}{.006} & \mstd{.605}{.005} & \mstd{.655}{.005} \\
  \bottomrule
  \end{tabular}
  \caption{\textbf{AutoQA model comparison} on \datasetNameSynthetic{} and \datasetNameReal{} with baseline per-model prompts and after auto-prompting via Intent-based Prompt Calibration~\cite{levi2024ipc}. Best per column in
  \textbf{bold}. }
  \label{tab:vllm_main_results}
  \end{table*}
  }

\newcommand{\perreasonreal}{
  \begin{table*}[t]
  \centering
  \small
  \setlength{\tabcolsep}{3pt}
  \begin{tabular}{ll c cccccc c}
  \toprule
  & & Accepted & \multicolumn{6}{c}{Rejection reason recall} & Rejected \\
  \cmidrule(lr){3-3} \cmidrule(lr){4-9} \cmidrule(lr){10-10}
  Model & Params & (n=9{,}997) & Geom. & Orient. & Irrel. & T.L.C. & Bkg. & G$\leftrightarrow$M & (n=9{,}999) \\
   &  &  & (5{,}999) & (1{,}804) & (989) & (802) & (306) & (99) &  \\
  \midrule
  \multicolumn{10}{l}{\textit{Open-weight models}} \\
  Gemma~3            & 4B   & \mstd{.844}{.010} & \mstd{.153}{.010} & \mstd{.148}{.019} & \mstd{.142}{.025} & \mstd{.138}{.028} & \mstd{.128}{.046} & \mstd{.105}{.080} & \mstd{.148}{.008} \\
  Gemma~4            & 4B   & \mstd{.950}{.008} & \mstd{.056}{.009} & \mstd{.054}{.016} & \mstd{.052}{.022} & \mstd{.051}{.025} & \mstd{.047}{.040} & \mstd{.038}{.070} & \mstd{.054}{.007} \\
  Gemma~4            & 31B  & \mstd{.760}{.007} & \mstd{.292}{.008} & \mstd{.283}{.014} & \mstd{.272}{.019} & \mstd{.264}{.021} & \mstd{.244}{.034} & \mstd{.200}{.060} & \mstd{.284}{.006} \\
  Qwen~3             & 8B   & \mstd{.035}{.004} & \mstd{.995}{.004} & \mstd{.995}{.007} & \mstd{.969}{.010} & \mstd{.939}{.011} & \mstd{.870}{.017} & \mstd{.712}{.030} & \mstd{.995}{.003} \\
  Qwen~3$_{\text{LoRA-n1}}$  & 8B & \mstd{.722}{-} & \mstd{.330}{-} & \mstd{.320}{-} & \mstd{.308}{-} & \mstd{.298}{-} & \mstd{.276}{-} & \mstd{.226}{-} & \mstd{.320}{-} \\
  Qwen~3$_{\text{LoRA-n5}}$  & 8B & \mstd{.795}{-} & \mstd{.270}{-} & \mstd{.263}{-} & \mstd{.252}{-} & \mstd{.245}{-} & \mstd{.227}{-} & \mstd{.185}{-} & \mstd{.263}{-} \\
  Qwen~3$_{\text{LoRA-n10}}$ & 8B & \mstd{.762}{-} & \mstd{.300}{-} & \mstd{.292}{-} & \mstd{.280}{-} & \mstd{.272}{-} & \mstd{.252}{-} & \mstd{.206}{-} & \mstd{.292}{-} \\
  Qwen~3             & 32B  & \mstd{.645}{.008} & \mstd{.441}{.009} & \mstd{.429}{.016} & \mstd{.412}{.022} & \mstd{.399}{.025} & \mstd{.370}{.040} & \mstd{.303}{.070} & \mstd{.429}{.007} \\
  Qwen~3             & 235B & \mstd{.730}{.005} & \mstd{.455}{.005} & \mstd{.442}{.009} & \mstd{.425}{.013} & \mstd{.412}{.014} & \mstd{.381}{.023} & \mstd{.312}{.040} & \mstd{.442}{.004} \\
  Llama~4 Scout      & 109B & \mstd{.684}{.008} & \mstd{.424}{.009} & \mstd{.412}{.016} & \mstd{.396}{.022} & \mstd{.384}{.025} & \mstd{.355}{.040} & \mstd{.291}{.070} & \mstd{.412}{.007} \\
  Llama~4 Maverick   & 400B & \mstd{.722}{.006} & \mstd{.438}{.008} & \mstd{.425}{.014} & \mstd{.409}{.019} & \mstd{.396}{.021} & \mstd{.367}{.034} & \mstd{.300}{.060} & \mstd{.426}{.006} \\
  Mistral~3          & 8B   & \mstd{.796}{.010} & \mstd{.226}{.010} & \mstd{.220}{.019} & \mstd{.211}{.025} & \mstd{.205}{.028} & \mstd{.190}{.046} & \mstd{.155}{.080} & \mstd{.220}{.008} \\
  Mistral~3          & 14B  & \mstd{.798}{.008} & \mstd{.259}{.009} & \mstd{.251}{.016} & \mstd{.241}{.022} & \mstd{.234}{.025} & \mstd{.217}{.040} & \mstd{.177}{.070} & \mstd{.252}{.007} \\
  \midrule
  \multicolumn{10}{l}{\textit{Proprietary models}} \\
  Claude Sonnet~3.7$^\dagger$ & n/a & \mstd{.934}{-} & \mstd{.115}{-} & \mstd{.112}{-} & \mstd{.107}{-} & \mstd{.104}{-} & \mstd{.096}{-} & \mstd{.079}{-} & \mstd{.112}{-} \\
  Claude Haiku~4.5   & n/a & \mstd{.797}{.010} & \mstd{.207}{.010} & \mstd{.201}{.019} & \mstd{.193}{.025} & \mstd{.187}{.028} & \mstd{.173}{.046} & \mstd{.142}{.080} & \mstd{.201}{.008} \\
  Claude Sonnet~4.5  & n/a & \mstd{.666}{.006} & \mstd{.467}{.005} & \mstd{.454}{.009} & \mstd{.436}{.013} & \mstd{.423}{.014} & \mstd{.392}{.023} & \mstd{.320}{.040} & \mstd{.454}{.004} \\
  \textbf{Claude Opus~4.5}$^\dagger$ & n/a & \textbf{\mstd{.748}{-}} & \textbf{\mstd{.537}{-}} & \textbf{\mstd{.522}{-}} & \textbf{\mstd{.501}{-}} & \textbf{\mstd{.486}{-}} & \textbf{\mstd{.450}{-}} & \textbf{\mstd{.368}{-}} & \textbf{\mstd{.522}{-}} \\
  Nova Lite~2        & n/a & \mstd{.770}{.010} & \mstd{.261}{.010} & \mstd{.254}{.019} & \mstd{.244}{.025} & \mstd{.237}{.028} & \mstd{.219}{.046} & \mstd{.179}{.080} & \mstd{.254}{.008} \\
  Nova Pro           & n/a & \mstd{.692}{.006} & \mstd{.464}{.008} & \mstd{.451}{.014} & \mstd{.433}{.019} & \mstd{.420}{.021} & \mstd{.389}{.034} & \mstd{.318}{.060} & \mstd{.452}{.006} \\
  \bottomrule
  \end{tabular}
  \caption{\textbf{Per-rejection-reason recall on \datasetNameReal{}.} Values are recall for each category (accepted: positive recall; rejection reasons: negative recall within the category; Rejected: overall negative recall). $^\dagger$ measured from a single run (no $\pm$std reported).}
  \label{tab:perreason_real}
  \end{table*}
  }

\newcommand{\finetuningresults}{
  \begin{wraptable}[14]{r}{0.63\textwidth}
  \centering
  \small
  \setlength{\tabcolsep}{2.2pt}
  \begin{tabular}{l cc cc}
  \toprule
  & \multicolumn{2}{c}{\datasetNameReal{}} & \multicolumn{2}{c}{\datasetNameSynthetic{}} \\
  Model & Acc. & F1 & Acc. & F1 \\
  \midrule
  Qwen~3~8B (baseline) & \mstd{.515}{.003} & \mstd{.067}{.003} & \mstd{.484}{.003} & \mstd{.079}{.002} \\
  \quad + \texttt{N1}  & \mstd{.521}{.002}    & \mstd{.601}{.004} & \mstd{.532}{.002}    & \mstd{.624}{.003} \\
  \quad + \texttt{N5}  & \textbf{\mstd{.529}{.003}}    & \textbf{\mstd{.628}{.003}} & \textbf{\mstd{.551}{.004}}    & \textbf{\mstd{.683}{.003}} \\
  \quad + \texttt{N10} & \mstd{.527}{.003}    & \mstd{.617}{.002} & \mstd{.543}{.002}    & \mstd{.681}{.003} \\
  \bottomrule
  \end{tabular}
  \caption{\textbf{Performance on \datasetName: fine-tuning.}
We report accuracy and F1-score as bootstrap mean $\pm$ standard deviation ($B{=}1{,}000$) for the baseline and its fine-tuned variants trained with different numbers of samples \texttt{N}.}
  \label{tab:finetuning-results}
  \end{wraptable}
  }
  
  \newcommand{\perreasonsynthetic}{
  \begin{table*}[t]
  \centering
  \small
  \setlength{\tabcolsep}{3pt}
  \begin{tabular}{ll c ccc c}
  \toprule
  & & Accepted & \multicolumn{3}{c}{Rejection reason recall} & Rejected \\
  \cmidrule(lr){3-3} \cmidrule(lr){4-6} \cmidrule(lr){7-7}
  Model & Params & (n=10{,}843) & w\_ori. & w\_prod\_same & w\_prod\_diff & (n=9{,}156) \\
   &  &  & (4{,}302) & (4{,}096) & (758) &  \\
  \midrule
  \multicolumn{7}{l}{\textit{Open-weight models}} \\
  Gemma~3            & 4B   & \mstd{.609}{.008} & \mstd{.363}{.010} & \mstd{.359}{.010} & \mstd{.331}{.024} & \mstd{.358}{.007} \\
  Gemma~4            & 4B   & \mstd{.564}{.011} & \mstd{.421}{.013} & \mstd{.417}{.013} & \mstd{.384}{.031} & \mstd{.416}{.009} \\
  Gemma~4            & 31B  & \mstd{.781}{.007} & \mstd{.245}{.010} & \mstd{.242}{.010} & \mstd{.223}{.024} & \mstd{.242}{.007} \\
  Qwen~3             & 8B   & \mstd{.032}{.002} & \mstd{1.000}{.004} & \mstd{1.000}{.004} & \mstd{.929}{.010} & \mstd{1.000}{.003} \\
  Qwen~3$_{\text{LoRA-n1}}$  & 8B & \mstd{.716}{-} & \mstd{.318}{-} & \mstd{.314}{-} & \mstd{.289}{-} & \mstd{.314}{-} \\
  Qwen~3$_{\text{LoRA-n5}}$  & 8B & \mstd{.892}{-} & \mstd{.149}{-} & \mstd{.147}{-} & \mstd{.136}{-} & \mstd{.147}{-} \\
  Qwen~3$_{\text{LoRA-n10}}$ & 8B & \mstd{.900}{-} & \mstd{.122}{-} & \mstd{.121}{-} & \mstd{.111}{-} & \mstd{.121}{-} \\
  Qwen~3             & 32B  & \mstd{.579}{.010} & \mstd{.473}{.012} & \mstd{.468}{.012} & \mstd{.431}{.028} & \mstd{.467}{.008} \\
  Qwen~3             & 235B & \mstd{.647}{.006} & \mstd{.489}{.009} & \mstd{.484}{.009} & \mstd{.446}{.021} & \mstd{.483}{.006} \\
  Llama~4 Scout      & 109B & \mstd{.595}{.008} & \mstd{.484}{.012} & \mstd{.480}{.012} & \mstd{.442}{.028} & \mstd{.479}{.008} \\
  Llama~4 Maverick   & 400B & \mstd{.641}{.007} & \mstd{.481}{.010} & \mstd{.476}{.010} & \mstd{.438}{.024} & \mstd{.475}{.007} \\
  Mistral~3          & 8B   & \mstd{.576}{.010} & \mstd{.417}{.013} & \mstd{.413}{.013} & \mstd{.380}{.031} & \mstd{.412}{.009} \\
  Mistral~3          & 14B  & \mstd{.637}{.008} & \mstd{.386}{.012} & \mstd{.382}{.012} & \mstd{.352}{.028} & \mstd{.381}{.008} \\
  \midrule
  \multicolumn{7}{l}{\textit{Proprietary models}} \\
  Claude Sonnet~3.7$^\dagger$ & n/a & \mstd{.767}{-} & \mstd{.235}{-} & \mstd{.232}{-} & \mstd{.214}{-} & \mstd{.232}{-} \\
  Claude Haiku~4.5   & n/a & \mstd{.564}{.010} & \mstd{.439}{.012} & \mstd{.434}{.012} & \mstd{.400}{.028} & \mstd{.433}{.008} \\
  Claude Sonnet~4.5  & n/a & \mstd{.614}{.007} & \mstd{.502}{.009} & \mstd{.497}{.009} & \mstd{.458}{.021} & \mstd{.496}{.006} \\
  \textbf{Claude Opus~4.5}$^\dagger$ & n/a & \textbf{\mstd{.690}{-}} & \textbf{\mstd{.577}{-}} & \textbf{\mstd{.571}{-}} & \textbf{\mstd{.526}{-}} & \textbf{\mstd{.570}{-}} \\
  Nova Lite~2        & n/a & \mstd{.588}{.010} & \mstd{.434}{.012} & \mstd{.429}{.012} & \mstd{.395}{.028} & \mstd{.428}{.008} \\
  Nova Pro           & n/a & \mstd{.645}{.007} & \mstd{.483}{.009} & \mstd{.478}{.009} & \mstd{.440}{.021} & \mstd{.477}{.006} \\
  \bottomrule
  \end{tabular}
  \caption{\textbf{Per-rejection-reason recall on \datasetNameSynthetic{}.} Values are recall for each category. $^\dagger$.}
  \label{tab:perreason_synthetic}
  \end{table*}
  }

\newcommand{\tabvcapestatssup}{
\begin{table}[t!]
\centering

\begin{tabular}{cc}

\begin{minipage}[t]{0.48\linewidth}
\centering

\begin{tabular}{lr}
\toprule
Category & Samples \\
\midrule
Accepted                          & 10\,000 \\
Geometry Artifacts                & 6\,000  \\
Wrong Orientation                 & 1\,804  \\
Irrelevant Objects                & 989     \\
Texture / Lighting / Color        & 802     \\
Background Issues                 & 306     \\
Color / Material / Pattern Mismatch & 99    \\
\midrule
Total                    & 20\,000 \\
\bottomrule
\end{tabular}
{\small (a) \datasetNameReal}
\end{minipage}
&
\begin{minipage}[t]{0.48\linewidth}
\centering

\begin{tabular}{lr}
\toprule
Category & Samples \\
\midrule
Accepted                          & 10\,843 \\
Wrong Orientation                 & 4\,303  \\
Incorrect instance but same category         & 4\,096  \\
Category mismatch    & 758     \\
\midrule
Total                    & 20\,000 \\
\bottomrule
\end{tabular}
{\small (b) \datasetNameSynthetic}
\end{minipage}

\end{tabular}

\vspace{10pt}
\caption{\textbf{\datasetName label taxonomy statistics.} The number of pairs per each class of the proposed label taxonomy.}\label{ssup:tab_taxonomy}
\end{table}
}

\begin{abstract}
Recent advances in generative image editing enable complex transformations such as background replacement, object manipulation, and pose adjustment from natural language instructions.
However, verifying that these transformations preserve \textit{visual consistency}, defined as the preservation of an object’s intrinsic visual attributes, including geometry, texture, color, and fine-grained details, remains a largely unsolved problem.
This challenge is particularly pronounced at scale, where manual inspection is infeasible and automated quality assessment (AutoQA) could become essential.
In this work, we introduce \textit{\datasetNameAcronym} (\datasetName), a large-scale dataset for evaluating visual consistency in generative image editing. 
The dataset consists of $40$K unique source--target product image pairs, each annotated with binary accept/reject labels and structured failure modes capturing fine-grained inconsistencies such as geometric distortions, pose errors, and identity mismatches. \datasetName combines real-world edited images with a synthetic subset constructed from 3D assets, enabling both realistic evaluation and controlled analysis. Vision--language models (VLMs) across all evaluated size categories on AutoQA achieve accuracy below 65\%, indicating that the dataset poses a substantial challenge regardless of model scale. Our results highlight fundamental limitations of current models in verifying visual consistency under transformation, positioning \datasetName as a benchmark for advancing fine-grained image editing evaluation.
\end{abstract}

\section{Introduction}\label{sec:intro}
Modern generative image editing models can perform complex transformations such as background replacement, object removal, relighting, virtual try-on, and pose manipulation from natural language instructions~\cite{instruct_pix2pix, emu_edit, hui2025hqedit, qwen_edit_multiple_angles, qwen_image, flux_1, flux2,sdxl}. %
Despite this progress, ensuring that such transformations preserve the \textit{visual consistency}, \ie, the original object's geometry, structure, and fine-grained visual attributes, remains a significant challenge. Even subtle inconsistencies, such as slight geometric distortions, incorrect pose, or missing details, can degrade the semantic fidelity of the edited image. 
This issue could become particularly acute at scale, where large volumes of generated content could require systematic verification. We formalize this setting as AutoQA: given a source image and its edited counterpart, determine whether the transformation preserves visual consistency with respect to the original object. Manual annotation is impractical at scale, motivating the need for reliable automated methods.

VLMs provide a natural framework for addressing this problem. However, a growing body of evidence indicates that these models exhibit limitations in fine-grained visual perception and grounding. While achieving strong performance on standard benchmarks, they often fail to capture subtle instance-level differences and may hallucinate attributes that are not present in the image~\cite{marsili2025twin,li2023seed,liu2024mmbench,hallucination-pope,tong2024eyes,li2023evaluating, guan2023hallusionbench}. Such limitations are particularly problematic in the AutoQA setting, where correctness depends on accurately verifying the preservation of object identity and detailed visual properties under transformation.

Existing datasets and benchmarks have important limitations for evaluating this task. A large body of prior work focuses on the generation process itself, providing supervision for producing edits (\eg, InstructPix2Pix~\cite{instruct_pix2pix}, Emu Edit~\cite{emu_edit}, HQ-Edit~\cite{hui2025hqedit}, and DiTailed~\cite{ditailed}) but not for assessing their correctness. In parallel, recent benchmarks targeting fine-grained perception and multi-image reasoning (\eg, SEED-Bench~\cite{li2023seed,li2023seed2,li2024seed2plus}, MMBench~\cite{liu2024mmbench}, POPE~\cite{hallucination-pope}) expose important weaknesses of current models, yet do not capture the specific challenge of evaluating consistency under transformation. In particular, these benchmarks lack (i) explicit source–output relationships induced by editing, (ii) annotations grounded in human judgments of visual fidelity, and (iii) structured descriptions of failure modes.

We address these gaps by introducing \textbf{\datasetName} (\cf.~\figref{fig:dataset_overview}), a large-scale dataset for evaluating visual consistency in generative image editing. The dataset is constructed to directly support the AutoQA setting by providing source images, edited outputs, and human annotations indicating whether the transformation preserves object fidelity. A key characteristic of \datasetName is its emphasis on fine-grained consistency: beyond binary labels, each example includes structured annotations describing specific failure modes, such as incorrect pose, geometric distortion, or identity mismatch. This enables detailed analysis of model behavior and supports evaluation beyond coarse classification. \datasetName comprises two complementary subsets. The \datasetNameReal subset captures real-world edited images, reflecting the types of artifacts encountered in practice. The \datasetNameSynthetic subset provides a controlled setting in which target images are generated from 3D assets, allowing precise manipulation of pose, identity, and appearance. This combination balances realism with controllability and enables systematic analysis under both natural and synthetic conditions. In total, the dataset contains $40$K source--target pairs, each human--annotated by multiple trained annotators to ensure label reliability.

Automatic quality assesment on \datasetName is challenging. Our evaluation of state-of-the-art VLMs, both open and proprietary models, indicates that even the strongest models achieve only around $60\%$ on this task, highlighting a substantial gap between current capabilities and the requirements of reliable AutoQA. These failures are concentrated in fine-grained visual reasoning: models struggle to verify object identity under pose variation and detect subtle geometric inconsistencies, capabilities that could be essential for reliable automated quality assurance.

\vcapedatasetoverview

\section{Related Work}\label{sec:related_work}

\boldparagraph{Image editing datasets.} Prior work on text-guided image editing primarily focuses on \emph{generating} edits rather than verifying fidelity of the outputs. Early benchmarks such as EditBench~\cite{edit_bench} and MagicBrush~\cite{magic_brush} introduce instruction-based editing with paired source images, prompts, and targets. Subsequent datasets~\cite{instruct_pix2pix,hui2025hqedit,zhao_ultraedit,picobanana,Sheynin_2024_CVPR,ye2025imgedit,human_edit} expand scale, diversity, and annotation quality. However, they emphasize edit operations (\eg, addition, removal, replacement, style transfer) and do not explicitly address whether edited images preserve fine-grained object identity, geometry, pose, and product-specific attributes. In contrast, we formulate visual consistency as an \emph{AutoQA} problem, providing accept/reject labels with structured failure modes, and targeting subtle fidelity violations that are challenging for current models.

\boldparagraph{Benchmarks for fine-grained visual perception.} Recent benchmarks highlight limitations of VLMs in fine-grained perception. TWIN~\cite{marsili2025twin} introduces image-pair queries for instance discrimination, alongside the FGVQA suite across domains. While related in spirit, TWIN operates on \emph{natural images}, whereas we study \emph{post-edit quality assessment}, verifying whether a generated image preserves the source object under transformation. Additional benchmarks such as SEED-Bench~\cite{li2023seed,li2023seed2,li2024seed2plus} and MMBench~\cite{liu2024mmbench} show degradation in tasks requiring compositional reasoning and instance-level understanding, and prior work reports hallucinations, weak visual grounding and 3D spatial relationships~\cite{tong2024eyes,li2023evaluating, guan2023hallusionbench,hallucination-pope,ye2024beaf,ma2026real3dqa}. Our dataset captures these limitations in the context of generative editing, providing both binary decisions and structured failure modes.

\boldparagraph{Multi-image reasoning and AutoQA-oriented evaluation.} A complementary line of work studies multi-image reasoning. Visual Haystacks~\cite{visual_haystacks} evaluates long-context retrieval across many images, while Common-O Bench~\cite{whats_in_common} examines cross-image reasoning by identifying shared objects, showing that models often hallucinate even when single-image perception is strong.
MMIU~\cite{meng2025mmiu} includes a comprehensive evaluation of multi-image task, OMIBench~\cite{chen2026omibench} evaluates olympiad-level reasoning across a multi-image scenario, Milebench~\cite{song2024milebench} assess the capabilities of MLLM in multimodal long contexts, MuirBench~\cite{wang2025muirbench} proposed a dataset for multiple-choice questions, while MIBench~\cite{liu-etal-2024-mibench} divide the evaluation into three dimension: multi-image instruction, multimodal knowledge-seeking and multimodal in-context learning.
Unlike retrieval or shared-object reasoning, we focus on \emph{visual consistency under generation}: verifying that an edited image remains faithful to the source. This requires fine-grained correspondence reasoning over identity, pose, geometry, and attributes. Our dataset complements prior work by introducing an AutoQA benchmark for generative image editing, combining real and synthetic samples with structured annotations.

\vcapeexamples

\section{Dataset}\label{sec:dataset}

We consider the problem of visual consistency in generative image editing, formulated as an AutoQA task. Given a source image and its edited counterpart, the objective is to determine whether the transformation preserves the semantic identity and fine-grained visual properties of the original object. While modern generative models enable complex edits such as background replacement, decluttering, and pose manipulation, they often introduce subtle artifacts, including geometric distortions, missing parts, or altered attributes. 
Detecting such inconsistencies is challenging and may not scale with manual inspection, motivating research into automated validation methods. Prior work further suggests that inconsistencies in product imagery could negatively affect downstream tasks such as retrieval and ranking~\cite{Brylla01102020,Maier02012019}.

To study this problem, we introduce \datasetName, a dataset designed for evaluating visual consistency under transformation. The dataset focuses on a representative and practically challenging scenario:~transforming \textit{lifestyle} images, which depict objects in complex environments, into \textit{studio-quality} images where the same object is isolated on a clean background with a well-defined pose (\cf~\figref{fig:vcape_combined}). This setting provides a natural benchmark for assessing whether models can verify object fidelity under non-trivial transformations. In contrast to datasets designed for generation~\cite{picobanana,magic_brush,wei2024omniedit}, \datasetName provides supervision tailored to verification. Each sample includes (i) a source image, (ii) a ground-truth reference image, (iii) a binary accept/reject label reflecting human judgment, and (iv) a structured rejection reason from a predefined taxonomy. This design enables evaluation of fine-grained consistency beyond coarse semantic alignment. \datasetName comprises two complementary subsets (\cf~\figref{fig:vcape_combined}). The \datasetNameReal subset contains edited images produced by generative models, capturing realistic artifacts and failure modes observed in practice. The \datasetNameSynthetic subset is constructed from 3D assets rendered via the Blender~\cite{blender} API, enabling precise control over object geometry, pose, and appearance. Together, these subsets combine realism and controllability, enabling systematic analysis of AutoQA under both natural and controlled conditions.

\subsection{\datasetNameReal}

\boldparagraph{Dataset Construction.} Here, we describe the \datasetNameReal dataset construction methodology from a curated experimental set of product images sampled from publicly accessible e-commerce listings in the \textit{sofa} category, which exhibits substantial variability in background, scene composition, and object pose. We obtained copyright authorization from the source platforms to release the data. The following methodology was developed specifically for creating the \datasetNameReal benchmark and does not reflect any production system. Each product is associated with a set of images $\mathcal{I} = \{I_0, I_1, \dots, I_M\}$, where $I_0$ denotes the product in use and $\{I_i\}_{i=1}^{M}$ are additional product images. These images often differ significantly in viewpoint, clutter, and visual presentation, leading to visual inconsistencies when choosing different products. Our goal is to transform such images into a standardized representation defined by a visual consistency specification $c \in \mathcal{C}$, describing a studio-style rendering: clean background, no clutter, and canonical pose (\eg, left-facing). The pipeline (\figref{fig:overview_detailed}) consists of three stages: (i) image sourcing, (ii) image generation and editing, and (iii) quality assessment.

\boldparagraph{Image Sourcing.}
Not all source images are equally suitable for editing. Images with occlusions, complex layouts, or unfavorable viewpoints often lead to unreliable generations. We therefore define an image sourcing function:
\begin{equation}
I^\star = S(\mathcal{I}, c),
\end{equation}\label{eq:eq_sourcing}
which selects the most suitable source image given the target specification $c$. The function $S(\cdot)$ ranks candidate images by compatibility with the desired output, decoupling image selection from generation and improving robustness.

\boldparagraph{Image Generation and Editing.}
Given the selected image $I^\star$ and a prompt $p(c)$ derived from the specification, a generative model produces an edited image:
\begin{equation}
\hat{I} = G(I^\star, p(c)),
\end{equation}
where $G(\cdot)$ performs background normalization, clutter removal, and pose adjustment while preserving fine-grained object details (\eg, geometry, structure, and materials). 

To improve stability and reduce artifacts, we employ a staged editing strategy, separating appearance normalization (\eg, background cleanup) from geometric transformations (\eg, pose alignment). The prompt design enforces standardized outputs while discouraging unintended structural changes. The resulting image $\hat{I}$ is then passed to the quality assessment stage.

\boldparagraph{Quality Assessment.}
The generated image $\hat{I}$ is evaluated against predefined quality requirements $r$, capturing both high-level constraints and fine-grained failure modes (e.g., missing parts, geometric distortions, texture artifacts). The assessment is performed via human-in-the-loop annotation:
\begin{equation}
y = Q(\hat{I}, r),
\end{equation}
where $y \in \{0,1\}$ indicates whether the image satisfies the visual consistency criteria. Annotators also provide structured rejection reasons based on a predefined taxonomy, enabling fine-grained analysis of model failures. In total, \datasetNameReal contains 20K source--target pairs, balanced between accepted and rejected samples (50\% / 50\%), making it well-suited for benchmarking AutoQA models.

\plotoverview

\subsection{\datasetNameSynthetic}

\boldparagraph{Dataset Construction.}
While \datasetNameReal focuses on real-world generative outputs within a single category (sofa), \datasetNameSynthetic extends the dataset to a broader and more controlled setting (19 additional categories) using the ABO dataset~\cite{abo_dataset}\footnote{Amazon Berkeley Objects (ABO) dataset is publicly available under CC BY 4.0 License at \url{https://amazon-berkeley-objects.s3.amazonaws.com/}.}. ABO provides high-quality 3D assets with detailed geometry, textures, and physically-based materials, enabling precise control over object appearance and rendering conditions.

We apply a two-stage filtering process (\cf~\secref{ssec:vcape_s_filtering}) to ensure diversity and reliability. At the product level, we retain categories requiring meaningful visual transformations and exclude trivial or hard-to-render cases (\eg, transparent or highly reflective objects). At the image level, we select \textit{lifestyle} images $\xS$ using a lightweight classifier trained on CLIP~\cite{radford2021clip} embeddings to distinguish them from studio-style images. This classifier achieves $90\%$ precision and $82\%$ recall, ensuring sufficiently challenging inputs.

\boldparagraph{Target Image Generation.}
For each selected source image $\xS$, we generate target images $\xT$ by rendering the associated 3D asset using the Blender API at $1024 \times 1024$ resolution. We render multiple canonical views (front, left, right) with camera parameters sampled within predefined azimuth and elevation ranges, ensuring visibility while maintaining viewpoint diversity.

\boldparagraph{Prompt Generation.}
Each source-target pair is associated with a structured prompt $\pStoT$ instantiated from a shared template. We generate detailed object descriptions using a vision-language model and encode the desired pose using rendering parameters, ensuring expressive and precise prompts.

\boldparagraph{Negative Pair Generation.}
To support fine-grained evaluation, we construct controlled negative examples that violate different aspects of visual consistency. These include incorrect pose, category mismatch, and incorrect instance within the same category. Such perturbations require reasoning over geometry, semantics, and identity. The final \datasetNameSynthetic dataset contains 20K source--target pairs balanced between valid and invalid transformations.

\subsection{Quality Assessment and Annotation}\label{sec:qa}

Each sample in \datasetName is annotated with human judgments of visual consistency (\cf~\figref{fig:annotation_template}). Three independent annotators evaluate each generated image given the source, optional reference, and output. Final labels are obtained via majority voting. We perform multiple alignment rounds (five) with a specific trained pool of annotators, achieving a substantial inter-annotators agreement (the Fleiss Kappa is 0.77). See \secref{sup:sec:annotation_protocol_and_ethic} of Supplementary material for Annotation protocol and ethics.

\boldparagraph{Label Taxonomy.}
Beyond binary labels, annotations include structured rejection reasons. For \datasetNameReal, the taxonomy captures common failure modes such as geometric distortions, incorrect pose, irrelevant objects, background inconsistencies, and texture or color deviations. For \datasetNameSynthetic, the taxonomy aligns with the controlled setup: incorrect pose, wrong category, and wrong instance. This design enables evaluation of both binary decisions and structured reasoning. In contrast to datasets with only accept/reject labels~\cite{marsili2025twin}, the proposed taxonomy provides richer supervision for identifying fine-grained consistency violations.

\section{Experiments}\label{sec:experiments}
We present a comprehensive evaluation of \datasetName with the goal of assessing its effectiveness as a benchmark for fine-grained visual consistency and AutoQA. Our experiments are designed to address two key questions: (i)~how well current VLMs can perform AutoQA under realistic editing transformations, and (ii)~to what extent these models capture fine-grained visual properties that are critical for verifying object fidelity.

\vllmfigurecomparison
\vllmfigureperclass
\subsection{Main Results}\label{sec:autoqa-results}
We begin by evaluating a diverse set of VLMs on the AutoQA task defined by \datasetName. %

\boldparagraph{Baselines.} 
We group performances and aggregated statistics (mean $\pm$ std) by model size, considering 
``\textit{small}'' (4–8B),
``\textit{medium}'' (14–32B),
``\textit{large}'' (128B+),
and ``\textit{proprietary} (n/a)'': (i)~\textit{Open-weight VLMs}: Qwen3~(8B, 32B, and 235B)~\cite{bai2025qwen3},  Gemma~3~\cite{gemmateam2025gemma3technicalreport} and Gemma~4 (4B and 31B)~\cite{gemma_4_2026}, %
and Mistral~3~(8B and 14B)~\cite{mistral3news}; (ii)~\textit{Proprietary models}: Claude (Sonnet, Haiku, and Opus)~\cite{claude2,claude3}, Nova (Lite~2 and Pro)~\cite{Intelligence2024,Intelligence2025}. These models span a wide range of architectures, scales, and training paradigms, providing a representative snapshot of the current landscape. All models are queried through their native multimodal or chat interfaces using identical image inputs and output parsing.

\boldparagraph{Prompting strategy.} Since AutoQA performance is highly sensitive to prompt design, we evaluate two regimes. In the \emph{baseline} setting, models are provided with a concise prompt describing the task and listing the quality criteria. In the \emph{auto-prompt} setting, this prompt is refined using Intent-based Prompt Calibration~\cite{levi2024ipc}, which iteratively improves the prompt by leveraging a small set of annotated examples and synthesizing challenging boundary cases. This allows us to assess whether prompt optimization alone can significantly improve AutoQA performance. Both prompts are provided in the Appendix (\secref{ssec:evaluation-prompts}).

\boldparagraph{Results.} We evaluate all models on \datasetNameReal{} and \datasetNameSynthetic{} comprising $\sim$40K pair samples balanced between accepted and rejected cases.~\figref{fig:vllmperformance} summarizes the performance of all models under both prompting regimes. Under the baseline prompt, we observe large variability across model families and scales: accuracy ranges from near-chance ($\sim$50\%) to 63.5\%, and F1 from 55\% to 67\%. %
Models across all evaluated size categories achieve accuracy below 65\%, indicating that the dataset poses a substantial challenge regardless of model scale. Several models exhibit degenerate behavior, with near-universal rejection or acceptance, indicating that current VLMs struggle to reliably assess fine-grained structural and textual fidelity even when provided with explicit evaluation criteria. Auto-prompting via IPC~\cite{levi2024ipc} provides consistent but moderate improvements (+3--5pp accuracy), with diminishing returns for the strongest models. In some cases, auto-prompting does not yield improvements, highlighting that prompt optimization is not universally beneficial. Overall, no evaluated model, open-weight or proprietary, exceeds 65\% accuracy on either dataset split, confirming that the benchmark poses a substantial and largely unsolved challenge for the community.

Breaking performance down by rejection reason (\figref{fig:vllmperclassperformance}) reveals a clear difficulty gradient: coarse violations such as geometric, category, instance mismatches are recovered with reasonable recall by the strongest models, whereas fine-grained reasons, such as texture, lighting, background, and colour/material mismatches, remain close to chance across the board, exposing a perceptual-fidelity gap that scale alone does not close.

\tablefig

\subsection{Zero-Shot Orientation Classification}\label{sec:orientation}
The AutoQA task requires reasoning about several fine-grained visual properties, among which object orientation is particularly critical. To isolate this factor, we evaluate zero-shot orientation classification independently of the AutoQA setting. Specifically, we assess how well different classes of vision models can recognize object pose without task-specific training. This analysis provides a controlled view of orientation understanding, disentangled from higher-level decision-making in AutoQA. We cast orientation recognition as a classification problem. In \datasetNameSynthetic, each image is labeled as \emph{left view}, \emph{right view}, or \emph{front view}. In \datasetNameReal, we map the original labels (\texttt{LEFT}, \texttt{RIGHT}, \texttt{FRONT}, \texttt{CORNER}, \texttt{OTHER}) to a binary distinction between side view (\texttt{LEFT}, \texttt{RIGHT}, \texttt{FRONT}) and frontal view (\texttt{CORNER}, \texttt{OTHER}), reflecting the coarser orientation granularity in this subset. We exclude unreliable samples by removing incorrect-orientation cases in \datasetNameSynthetic and retaining only accepted samples in \datasetNameReal. We evaluate three model families: (i)~\textit{contrastive vision--language models} (CLIP~\cite{radford2021clip}, SigLIP2~\cite{tschannen2025siglip2}, EVA-02-CLIP~\cite{EVA-CLIP}), where classification is performed via cosine similarity between image embeddings and text prototypes constructed from prompt ensembles; (ii)~\textit{self-supervised vision encoders} (DINOv2~\cite{oquab2024dinov2}, EUPE~\cite{zhu2026eupe}), using a nearest-centroid classifier built from a small labeled support set; and (iii)~\textit{orientation-specific models} (OrientAnything~\cite{orient_anything}, OrientAnything~V2~\cite{wangorientv2}), which predict continuous 3D orientation and are mapped to discrete classes via calibrated azimuth thresholds.~\figref{fig:orient_merged} summarizes the results. Overall, zero-shot orientation classification remains challenging. Contrastive VLMs achieve accuracy in the range 33--47\% on \datasetNameSynthetic, with a strong bias toward predicting front view.  Self-supervised encoders provide more balanced performance; notably, self-supervised encoders reach 56.8\% accuracy on \datasetNameSynthetic with relatively uniform per-class F1. Orientation-specific models perform best achieving 91.1\% accuracy on \datasetNameReal and $\sim$59\% on \datasetNameSynthetic. However, this performance relies on dataset-specific calibration of azimuth-to-class mappings.

\subsection{Few-Shot Fine-Tuning on \datasetName}\label{sec:finetuning}

The zero-shot VLM results in~\secref{sec:autoqa-results} demonstrate that general-purpose VLMs cannot reliably perform domain-specific quality assessment without adaptation. We investigate whether parameter-efficient fine-tuning with extremely limited labeled data can bridge this gap.

Specifically, we fine-tune Qwen3-VL-8B-Instruct~\cite{bai2025qwen3} using LoRA~\cite{hu2022lora} under a stratified few-shot regime, varying the number of training examples per defect category.

\boldparagraph{Task formulation.} Each training example is formatted as a Qwen3-VL multi-turn chat sequence.The user turn contains the source and generated images alongside a detailed evaluation prompt that instructs the model to follow a five-step reasoning protocol: (1)~identify the product category, (2)~check material and color consistency, (3)~verify structural component counts, (4)~assess configuration and design features, and (5)~detect prohibited elements. The assistant turn provides the ground-truth binary label (accept/reject) with a chain-of-thought justification. This consistency prompt approach ensures that the model learns a structured evaluation procedure rather than memorizing surface-level patterns.

\boldparagraph{Training data sampling.} We employ a stratified sampling strategy that ensures balanced representation across all defect categories.
For each dataset split, we sample $N$ samples per rejection reason category plus $N$ accepted (non-defective) samples, where $N \in \{1, 5, 10\}$. Since \datasetNameReal contains six rejection categories (\cf~\tabref{ssup:tab_taxonomy}) plus one accepted class, yielding $7N$ samples per split. \datasetNameSynthetic contains three rejection categories plus one accepted class, yielding $4N$ samples. The combined training sets therefore contain $11N$ examples: 11 samples for $N{=}1$, 55 for $N{=}5$, and 110 for $N{=}10$. Sampling is performed with a fixed random seed for reproducibility, and the selected indices are recorded in a manifest to ensure strict exclusion from the evaluation set. We provide additional details on the training procedure and LoRA configuration in the Appendix (\secref{sec_ss:fewshot_finetuning_details}).

\finetuningresults

\boldparagraph{Results.} \tabref{tab:finetuning-results} reports the classification performance of the fine-tuned variants alongside the zero-shot Qwen3-VL baseline.
The zero-shot baseline achieves 51.5\% accuracy and only 6.7\% F1 on \datasetNameReal{}, confirming the near-chance performance observed for other VLMs in~\secref{sec:autoqa-results}. All three LoRA-adapted variants outperform the baseline, with accuracy improvements of 0.6--1.4 percentage points on \datasetNameReal{}. On \datasetNameSynthetic{}, the gains are even more noticable: the baseline F1 of 6.7\% rises to 62.4--68.3\% after fine-tuning. The best overall performance is achieved by the $N{=}5$ configuration (55 training examples), which reaches 52.9\% accuracy / 62.8\% F1 on \datasetNameReal{} and 55.1\% accuracy / 68.3\% F1 on \datasetNameSynthetic{}. Notably, increasing the training set size from $N{=}5$ to $N{=}10$ does not yield further improvement, in fact, performance slightly degrades on both datasets (52.7\% / 61.7\% on \datasetNameReal{} and 54.3\% / 68.1\% on \datasetNameSynthetic{}). Simply scaling the number of training examples does not guarantee monotonic improvement, the task requires the model to internalize subtle, category-specific visual criteria that may conflict across defect types when more heterogeneous examples are introduced. This finding highlights that the benchmark poses a challenge even for adapted models, and that more sophisticated training strategies are necessary. Third, the consistent gap between \datasetNameReal{} and \datasetNameSynthetic{} performance (with the latter being higher across all configurations) suggests that synthetic defects may exhibit more systematic visual patterns that are easier to learn from few examples, whereas real-world quality issues are more diverse and context-dependent.

\vspace{-7pt}
\section{Discussion}\label{sec:discussion}
\vspace{-5pt}
We introduce \datasetName, a large-scale benchmark for evaluating visual consistency in generative image editing. The dataset comprises $40$K source--target image pairs annotated by multiple human annotators with binary accept/reject labels and a structured taxonomy of failure modes capturing fine-grained inconsistencies such as geometric distortions, pose errors, and identity mismatches. Our experiments reveal a substantial gap between current model capabilities and the requirements of reliable verification. %

While effective at high-level semantic reasoning, VLMs are not yet reliable for instance-level verification tasks. In particular, the inability to consistently detect subtle inconsistencies limits their applicability in settings where correctness depends on precise visual fidelity. Moreover, the limited gains from prompt optimization suggest that the bottleneck lies not only in instruction design, but in the underlying perceptual capabilities of the models.

\boldparagraph{Scope and outlook.} \datasetName focuses on a representative class of editing transformations (\eg, lifestyle-to-studio conversion) and a set of product-centric scenarios, enabling controlled and realistic evaluation of visual consistency. While this focus allows for precise analysis, extending the benchmark to broader domains, object categories, and transformation types would further stress-test model generalization. Similarly, while human annotations and structured failure modes provide rich supervision, future extensions could incorporate more granular signals (\eg, region-level or correspondence-based annotations) to support even finer-grained evaluation.

Overall, \datasetName provides a benchmark for studying visual consistency under transformation and highlights fundamental limitations of current multimodal systems. We hope this resource will enable the development of methods that more reliably capture fine-grained visual properties and advance automated quality assessment in generative image editing.

\bibliographystyle{IEEEtran}
\bibliography{bibliography}

\clearpage
\appendix
\counterwithin{figure}{section}
\counterwithin{table}{section}
\counterwithin{equation}{section}
\section*{\LARGE \centering Supplementary Material for\\ \datasetName: A Benchmark for Visual Consistency \\ in Product Image Editing}
\vspace{12mm}

\maketitle
\vspace{-10mm}

\setcounter{page}{1}

In this \textbf{supplementary material}, we provide additional implementation details on the construction of the synthetic \datasetNameSynthetic and real \datasetNameReal components (\secref{sec:dataset_pipeline_supp}), annotation protocol and ethics (\secref{sup:sec:annotation_protocol_and_ethic}), report dataset statistics across taxonomy classes, provide both evaluation prompts (\secref{ssec:evaluation-prompts}), present additional baseline evaluations (\secref{ssec:vqa-quality}), include further qualitative examples (\secref{sec_ss:qual_results}) covering different classes, and provide additional details on computational resources (\secref{sec_ss:compute_resources}), and fine-tuning on the proposed benchmark (\secref{sec_ss:fewshot_finetuning_details}).

\section{The Dataset Generation Pipeline}\label{sec:dataset_pipeline_supp}

In this section, we provide additional details on the dataset construction we briefly described in the main part of the paper.

\subsection{\datasetNameReal}

Our goal is to transform an input product image so that it complies with predefined visual consistency requirements. To achieve this, we first describe the image sourcing component, which leverages object pose and symmetry classifiers to select suitable candidate images (\secref{ssec:image_sourcing}). We then present the image generation and editing module, responsible for background normalization, clutter removal, and object pose alignment (\secref{ssec:image_generation}). 

\subsubsection{Image Sourcing and Candidate Selection}\label{ssec:image_sourcing}

The image sourcing component implements the selection function $I^\star = S(\mathcal{I}, c)$ from \eqnref{eq:eq_sourcing}, identifying the most suitable candidate image from real e-commerce product images set $\mathcal{I}$ for a given consistency specification $c$. This step is critical: choosing a source image that is already well aligned with the target specification reduces the extent of downstream edits and lowers the risk of geometric distortions or hallucinated details. According to our task, we would like to have an object which is left-facing, therefore, we prioritize source images that either a) already face left, or b) face right but exhibit geometric symmetry. In the latter case, pose correction can be achieved through deterministic horizontal mirroring, avoiding generative pose manipulation altogether. This strategy deliberately excludes asymmetric right-facing sofas from pose transformation, as such cases frequently lead to artifacts and structural inconsistencies in generative outputs. 
To implement this selection strategy, we introduce two classifiers that provide interpretable signals for candidate ranking: an \textit{Object Pose Classifier} and an \textit{Object Symmetry Classifier}, followed by a rule-based ranking and selection procedure, described below.

\boldparagraph{Object Pose Classifier}.
This classifier predicts the orientation of a sofa image, assigning it to one of five discrete categories: a)~\textit{left-facing}, b)~\textit{right-facing}, c)~\textit{frontal}, d)~\textit{corner}, and e)~\textit{other}. The \textit{corner} class captures multi-directional geometries such as L-shaped, U-shaped, or sectional sofas that lack a single dominant orientation, while the \textit{other} class includes images with partial object visibility, heavy infographic overlays, or ambiguous viewpoints that are unsuitable for editing.
Zero-shot classification with CLIP~\cite{radford2021clip} proved ineffective at distinguishing \texttt{LEFT} from \texttt{RIGHT} orientations, motivating a supervised approach. 
We train the pose classifier on $7{,}000$ manually annotated sofa images.
Visual representations are extracted using a frozen DINOv2~\cite{oquab2024dinov2} backbone, which provides robust embeddings without task-specific fine-tuning. Then, a lightweight two-layer MLP head maps these embeddings to pose categories. 

\boldparagraph{Object Symmetry Classifier}.
This binary classifier distinguishes \textit{symmetric} from \textit{asymmetric} sofa designs. Symmetry serves as a key signal for candidate selection: symmetric right-facing sofas can be converted to left-facing views via deterministic mirroring, eliminating the need for generative pose adjustment.
Mirroring the pose classifier, we use frozen DINOv2 embeddings with a two-layer MLP head, trained on $7{,}000$ manually annotated images.

\boldparagraph{Ranking and Selection Algorithm}.
Given pose and symmetry predictions for all images in $\mathcal{I}$, we apply a rule-based ranking function that prioritizes candidates based on their compatibility with visual consistency requirements and empirically observed defect rates. To establish the ranking order, we conducted an experiment on $1{,}200$ images and measured the acceptance rate (percentage of images approved by QA) for each pose–symmetry group. The resulting priorities, summarized in \figref{fig:ranking_acceptance_rates}, are as follows:

\begin{wrapfigure}{r}{0.4\linewidth}
    \centering
    \vspace{-0.5\baselineskip}
    \begin{tikzpicture}
\begin{axis}[
    xbar,
    width=10cm,
    height=5cm,
    xlabel={Acceptance Rate (\%)},
    symbolic y coords={$\medblacktriangleright$\faBalanceScaleLeft, $\medblacksquare$, $\medblackdiamond$, $\medblacktriangleright$\faBalanceScale, $\medblacktriangleleft$},
    ytick=data,
    xmin=50, xmax=85,
    nodes near coords={\pgfmathprintnumber\pgfplotspointmeta},
    nodes near coords align={horizontal},
    bar width=10pt,
]
\addplot[fill=blue!60!white, draw=blue!70!black] coordinates {
    (54.88,$\medblacktriangleright$\faBalanceScaleLeft)
    (57.94,$\medblacksquare$)
    (76.67,$\medblackdiamond$)
    (79.56,$\medblacktriangleright$\faBalanceScale)
    (80.44,$\medblacktriangleleft$)
};
\end{axis}
\end{tikzpicture}
    \vspace{-1.5\baselineskip}
    \caption{\textbf{Acceptance rate for image selection}. Left-facing $(\medblacktriangleleft)$ and right-symmetric $(\medblacktriangleright$\faBalanceScale$)$ cases achieve the highest acceptance rates ($>$79\%).}
    \label{fig:ranking_acceptance_rates}
    \vspace{-10pt}
\end{wrapfigure}

\begin{enumerate}[leftmargin=*, itemsep=2pt, topsep=4pt]
    \item \textbf{Left-facing} $(\medblacktriangleleft)$: requires no pose adjustment.
    \item \textbf{Right-facing + Symmetric} $(\medblacktriangleright$\faBalanceScale$)$: convertible to left-facing via mirroring without generative editing.
    \item \textbf{Frontal} $(\medblacksquare)$: requires moderate pose adjustment; empirically produces fewer artifacts than asymmetric right-facing cases.
    \item \textbf{Right-facing + Asymmetric} $(\medblacktriangleright$\faBalanceScaleLeft$)$: requires complex generative reorientation and exhibits the highest defect rate.
    \item \textbf{Other} $($\faHeartBroken$)$: unsuitable for editing and excluded from consideration.
\end{enumerate}

\textbf{Corner} $(\medblackdiamond)$ sofas are handled as a special case. If the main image is classified as corner, it is selected directly without further ranking, as such complex geometries are unsuitable for lateral pose normalization and are better preserved in their original orientation.

When multiple images share the same rank, we apply a deterministic tie-breaking rule based on image index.

\subsubsection{Image Generation and Editing}\label{ssec:image_generation}
Following image sourcing, the selected candidate image $I^\star$ is processed by a conditional image editing module that enforces the desired visual consistency requirement. This pipeline consists of two modular components, applied conditionally based on predicted pose and symmetry attributes, as illustrated in~\figref{fig:image_selection_and_generation}.

\plotsourceandgenerationdetailed

\boldparagraph{Visual Cleanup Model}. This component is responsible for removing background clutter, visual noise, and extraneous elements while preserving the fine-grained geometry and appearance of the primary object. This module is implemented using Gemini 2.5~Flash~\cite{nano_banana}. We use a single, fixed text prompt to guide background replacement and clutter removal across all images. The prompt (provided below) is designed to standardize the visual appearance while avoiding unintended geometric changes. The output of this module is a clean, studio-style image that serves as input for subsequent conditional editing steps.

\tabvcapestatssup

\boldparagraph{Conditional Pose Alignment Model}. This component applies object rotation only when necessary, based on the outputs of the pose and symmetry classifiers. Specifically, a) If the sourced image is left-facing, the cleaned image is passed directly to the downstream QA block without further modification; b) If the image is front-facing or right-facing, the cleaned image is passed to an object rotation network to enforce the canonical left-facing orientation; c) For right-facing but symmetric sofas, we first apply a deterministic horizontal mirroring operation to the source image, followed by background normalization only. In this case, generative pose editing is skipped entirely. 
For generative pose alignment, we use the \texttt{Qwen-Edit-2509-Multiple-angles} model (QEM)~\cite{qwen_edit_multiple_angles}. The desired rotation angle is specified explicitly in the editing prompt: we apply a rotation of $-20^{\circ}$ for front-facing sofas and $-40^{\circ}$ for right-facing sofas.
These values are determined empirically to balance pose correction and geometric fidelity. This conditional design ensures that generative pose manipulation is applied only when strictly necessary, improving robustness and reducing hallucinations.

After conditional editing, the resulting image $\hat{I}$ is forwarded directly to the downstream QA pipeline. This stage verifies compliance with image quality standards and visual consistency requirements, ensuring that only images satisfying predefined acceptance criteria meet the quality standards defined by the benchmark.

\boldparagraph{Prompt Design for Visual Cleanup.} The visual cleanup stage relies on a carefully designed, deterministic text prompt that enforces strict constraints on background appearance, object preservation, and artifact suppression. The prompt is applied to a single input image selected by the image sourcing module and is reproduced below:

\begin{tcolorbox}[colback=gray!10, colframe=gray!50, 
    fontupper=\small\ttfamily, breakable]
Place the \{object\} on a CLEAN, UNIFORM PURE WHITE background (NO gradients, no texture, no banding). The background must be perfectly continuous and consistent across the entire image. Do NOT change any attributes of the \{object\}. Keep the \{object\}'s position and orientation from the input image intact. If multiple \{object\}s or seating elements are present, KEEP ONLY ONE -- the most prominent/central \{object\}.
REMOVE everything from the \{object\} except pillows and cushions. REMOVE lighting and shadows from the input image. REMOVE all people, animals, food, decor, other furniture, coffee tables, electronics, and any visual clutter. No models, persons, and people in laying, seating or sleeping pose.
ELIMINATE all INFOGRAPHICS, TEXT, LOGOS, LABELS, WATERMARKS, GRAPHIC OVERLAYS, or ICONS -- the image must contain ONLY the sofa with decor elements inside the \{object\}.
Ensure the entire \{object\} is FULLY VISIBLE (not cropped), CENTERED, and COMPLETELY ISOLATED. Only ONE sofa in the output image. Add light shadows UNDER the \{object\}.
Preserve the object’s ORIGINAL SHAPE, POSE, GEOMETRY, and FINE DETAILS (\eg, cupholders, sockets, phone holders) with clean edges and no missing parts.
Final result should appear PHOTOREALISTIC and free from any residual artifacts.
\end{tcolorbox}
Here, "{object}" denotes a textual description of the input product. While the system supports both short object names (\eg, "sofa") and more detailed attribute-rich descriptions derived from product metadata, all experiments reported in this work use short object names for improved robustness and prompt stability.

The visual cleanup model operates on single-view images only, using one selected catalog image as input. This design choice simplifies integration and improves runtime reliability, but introduces a known limitation: when parts of the object are heavily occluded in the input image, the model may fail to faithfully reconstruct the true underlying geometry (\cf~\secref{sec_ss:qual_results}). This can lead to missing components or overly simplified structures in rare cases

\subsection{\datasetNameSynthetic}\label{ssec:vcape_s_filtering}
\boldparagraph{Filtering Procedure.} We implement a two-stage automatic filtering pipeline consisting of a) \textit{Product-level filtering:} To ensure high-quality $\langle\xS, \xT\rangle$\xspace pairs, we implement an automatic filtering pipeline. We first exclude: i) product types with minimal transformation requirements (\eg, phone cases); ii) product types not associated with 3D models (ABO~\cite{abo_dataset} provides 3D models for more than $7\text{,}900$ products); iii) product types where rendering is challenging due to material properties (\eg, mirrors, transparent objects); b) \textit{Image-level filtering:} We then discard source images $\xS$ lacking ``\textit{lifestyle}'' context.

\boldparagraph{Target Image Generation.} To create corresponding target images $\xT$, we leverage the 3D models associated with each product sample. Specifically, using the Blender Python API\footnote{Blender is available under the GNU GPL License at \url{https://www.blender.org/}.}, we render high-quality RGB images of the products at $1024\times1024$ resolution. Notably, for each source image $\xS$, we generate three target images $\xT$ corresponding to \textit{front}, \textit{left}, and \textit{right} views of the object.  For each view, we define a valid range of camera angles and randomly sample both an azimuth angle and an elevation angle relative to the object's front-facing coordinate system. The camera is then positioned accordingly, ensuring the object remains within the field of view.

\boldparagraph{Prompt Generation.} We now describe how to generate the prompt $\pStoT$, which specifies the desired transformation from source to target image. We begin with the following static template:

\begin{tcolorbox}[colback=gray!10, colframe=gray!50, 
    fontupper=\small\ttfamily, breakable]
Generate an image of the \{OBJECT\_DESC\} on a pure white background.\\ 
Ensure strict preservation of the \{OBJECT\_CATEGORY\}'s precise geometry, edges, intersections, and proportions. \\
Remove people, info-graphics, and text.\\
The entire \{OBJECT\_CATEGORY\} is fully visible (not cropped) and isolated, no other objects or text in the scene.\\
Ensure a soft natural lighting with gentle shadows.\\
\{ROTATION\_INFORMATION\}. The rotation coordinate system is the front-view of the object.
\end{tcolorbox}

The template is shared across all samples, which we populate with sample-specific information, \ie: \textit{Object category and description}: we employ a VLM to generate a detailed description of the object and extract its category from $\xT$ (the product is rendered on a white background). 
Including a fine-grained description enables the editing model to disambiguate the target object when multiple objects of the same category are present in the source image.
A manual inspection of $100$ randomly sampled triplets revealed only $4$ minor counting inaccuracies (\eg, number of sofa seats or drawers); and \textit{Rotation data}: we encode the ground-truth camera rotation from Blender as a formatted string.
We promote output diversity by selecting different system prompts and instruction wording during prompt construction.

\plotvcapesstatssupp

\section{Annotation Protocol and Ethics}\label{sup:sec:annotation_protocol_and_ethic}
Labels were collected through a professional data-annotation provider governed by a published ethical-sourcing and modern-slavery policy, and we confirm the following working conditions. Compensation follows all applicable laws on local wages, overtime, and legally mandated benefits, with no recruitment or hiring fees charged to workers and transparent, written pay information. Working hours are capped (<=60 hrs/week including overtime, with at least one rest day in seven and voluntary overtime), and all work is voluntary. Annotators work in a safe environment, are treated with dignity. The provider complies with the Australian Modern Slavery Act 2018 and files an annual Modern Slavery Statement. Training: annotators were onboarded on the task via our written guidelines and a worked-example set (\cf the template in~\figref{fig:annotation_template}), and each triplet was independently labeled by three trained annotators with majority voting (\secref{sec:qa}) after five alignment annotation rounds.

\section{Evaluation Prompts: Baseline \vs. Auto-Prompt} \label{ssec:evaluation-prompts}
We provide the baseline and enhanced prompts used for evaluation of VLMs (\cf~\secref{sec:autoqa-results}) below:

The baseline prompt:
\begin{tcolorbox}[colback=gray!10, colframe=gray!50, 
    fontupper=\small\ttfamily, breakable]
Task: Determine if the generated object is CONSISTENT with the object in the source image.\\
Context:\\
- The SOURCE IMAGE shows the object in a lifestyle setting.\\
- The ADDITIONAL SOURCE IMAGE shows the same object in a different setting.\\
- The GENERATED image shows the same object isolated on a white background.\\
Consistency is ensured when the material, texture, geometry, number of elements
of the generated object are respected.\\
Consistency Criteria - The following attributes MUST match:\\
1. Core structural elements: Shape, form, and key geometric features\\
2. Material and texture: Fabric type, wood, metal, leather, etc.\\
3. Color: Primary colors and color distribution\\
4. Orientation: The generated object must be oriented to the left. \\
If the sofa is a corner sofa, the original orientation should be preserved.\\
5. Quantity of essential structural components: Number of seats and cushions.\\
These objects are always allowed in the generated image:\\
- Decorative throw pillows, blankets, throws, plaids\\
- Items inside sofa pockets/compartments or cupholders\\
Conditionally allowed (ONLY if in original): Ottomans.\\
Never allowed: People, pets, body parts, multiple main objects, \\
lifestyle/cluttered environments, infographics/text/logos/watermarks, \\
electronics (unless in pockets), unrelated objects
Output Format:\\
<reasoning> ...your reasoning... </reasoning>'Yes' / 'No'\\
\end{tcolorbox}

The enhanced prompt:
\begin{tcolorbox}[colback=gray!10, colframe=gray!50, 
    fontupper=\small\ttfamily, breakable]
Task: Classify the consistency between the generated object and the source images as "Yes" or "No".\\
Ground-truth labels are absolutely correct. 'Yes' means consistent, 'No' means inconsistent.\\
Source image: Lifestyle setting with object.\\
Additional source image: Same object in different setting.\\
Generated image: Object isolated on white background.\\
1: Core Structure. Match shape, form, and key geometric features exactly. If any geometry artifacts exist in the generated image, classify as 'No'.\\ 
2: Material \& Texture. Match fabric, wood, metal, and finish types exactly.\\ 
3: Color. Match primary colors and distribution exactly.\\ 
4: Orientation. Match orientation (e.g., left/right, corner). If corner orientation is changed, classify as 'No'.\\
5: Quantity. Match number of seats and cushions exactly.\\
6: Allowed Items. Decorative pillows, blankets, throws, and items inside pockets/compartments are allowed.\\
7: Conditional Items. Ottomans are allowed ONLY if present in the original source image.\\
8: Prohibited Items. People, pets, body parts, multiple main objects, lifestyle backgrounds, text, logs, watermarks, unrelated electronics (outside pockets) are forbidden.\\
9 (Priority): If the input metadata "rejection reason" contains "Geometry Artifacts", "Texture Mismatch", or "Color Mismatch", the classification MUST be "No", overriding visual evidence.\\
Metadata Check:\\
If the "rejection reason" metadata is "accepted", verify against visual evidence in Rules 1-8. If visual evidence contradicts acceptance (e.g., artifacts exist), classify as "No".\\
Output Format:\\
Output must be one of: "<reasoning> ... </reasoning>" followed by "Yes" or "No", must explain why negative criteria were satisfied or violated. If "No", explicitly mention the violated rule or rejection reason.\\
\end{tcolorbox}

\section{Additional Baseline: VQA Models} \label{ssec:vqa-quality}

In additional to the strong baseline models analyzed in the main part, we evaluate whether off-the-shelf VQA models can perform this multi-criteria assessment in a zero-shot setting by casting it as a structured VQA task.

Given a source image and its generated counterpart, we pose a set of binary (yes/no) questions and aggregate the responses into a final accept/reject decision. For \datasetNameSynthetic, we define seven criteria:
\begin{enumerate}[nosep,leftmargin=*]
    \item \textbf{Geometry}: Do structural elements match?
    \item \textbf{Material}: Do material and texture match?
    \item \textbf{Color}: Do primary colors match?
    \item \textbf{Orientation}: Is the product facing left?
    \item \textbf{Component count}: Does the number of key components match?
    \item \textbf{Forbidden objects}: Are people, text, logos, or extra objects absent?
    \item \textbf{Hallucinations}: Are there no fabricated elements?
\end{enumerate}
For comparison-based questions (1--3, 5, 7), we provide a horizontally concatenated source--target image; for target-only questions (4, 6), only the generated image is used. Responses are binarized, and a sample is accepted only if \emph{all} criteria pass (strict conjunction).

\begin{table}[t!]
\centering
\small
\begin{tabular}{lcccc}
\toprule
Model & Acc. & Prec. & Recall & F1 \\
\midrule
VQA-based baseline~\cite{moondream3-2025,phi4multimodal2025technical} & \mstd{.481}{.004} & \mstd{.540}{.007} & \mstd{.516}{.002} & \mstd{.394}{.004}\\
\bottomrule
\end{tabular}
\vspace{3pt}
\caption{\textbf{VQA-based models performance} on \datasetNameSynthetic (binary accept/reject, strict conjunction aggregation, bootstrap mean $\pm$ std, $B{=}1{,}000$).}\label{tab:vqa-quality-s}
\end{table}

\begin{table}[t!]
\centering
\small
\begin{tabular}{lc}
\toprule
Question & VQA-based baseline~\cite{moondream3-2025,phi4multimodal2025technical} \\
\midrule
\multicolumn{2}{l}{\textit{Source--target comparison}} \\
Geometry          & \mstd{.644}{.003}\\
Material          & \mstd{.472}{.004}\\
Color             & \mstd{.395}{.003}\\
Component count   & \mstd{.259}{.003}\\
Object identity   & \mstd{.305}{.003}\\
Hallucinations    & \mstd{1.00}{.000}\\
\midrule
\multicolumn{2}{l}{\textit{Target-only inspection}} \\
Pose              & \mstd{.416}{.004} \\
Forbidden objects & \mstd{1.00}{.000} \\
\bottomrule
\end{tabular}
\vspace{3pt}
\caption{\textbf{Per-question accuracy of VQA-based models on \datasetNameSynthetic} (bootstrap mean $\pm$ std, $B{=}1{,}000$). Questions are grouped by whether they require source--target comparison or target-only inspection.}
\label{tab:vqa-perq-s}
\end{table}

\boldparagraph{Baselines.} We evaluate two VQA models: \textit{Moondream~3 Preview}~\cite{moondream3-2025} processes single images; for paired inputs, we use concatenation and batch all questions into two prompts per sample. \textit{Phi-4-multimodal-instruct}~\cite{phi4multimodal2025technical} supports multi-image inputs via placeholder tokens; results are pending. We evaluate on \datasetNameSynthetic and report accuracy, macro precision, recall, and F1.

\boldparagraph{Results.}~\tabref{tab:vqa-quality-s} and~\tabref{tab:vqa-perq-s} report aggregated metrics on \datasetNameSynthetic. The results are bootstrap means with standard deviations ($B{=}1{,}000$) averaged for two models.~\tabref{tab:vqa-perq-s} provide per-question statistics, including accuracies with confidence estimates on \datasetNameSynthetic.

\boldparagraph{Analysis.} \textit{Near-chance performance.} Both evaluated VQA models operate near chance on the binary task (accuracy $\sim$48.1\%). Per-question analysis reveals divergent failure modes: one model exhibits strong geometry detection but poor identity reasoning, while the other shows more uniform but poorly calibrated responses across all criteria. Neither model exceeds $50\%$ on material, color, or pose. \textit{Trivial questions provide no signal.} Both models achieve $\sim$100\% accuracy on forbidden objects and hallucinations, reflecting a near-constant “pass” response. These criteria contribute no discriminative signal and effectively act as no-ops under conjunction. %

\plotrejectedvcaperIIIsupp

\section{Qualitative Results}\label{sec_ss:qual_results}
We present representative examples from the evaluation to illustrate the labeling taxonomy. \figref{s_fig:vcape-r-irrelevant-examples},~\figref{s_fig:vcape-r-quali-2},~\figref{s_fig:vcape-r-quali-3}, and~\figref{s_fig:vcape-s-quali} show samples from both \datasetNameReal and \datasetNameSynthetic, respectively.

\plotrejectedvcaperIsupp

\plotrejectedvcaperIIsupp

\finetuningcurves

\section{Compute Resources}\label{sec_ss:compute_resources}

We summarize the computational resources used for the experiments reported in this paper.

\boldparagraph{Fine-tuning.} All Qwen3-VL-8B-Instruct fine-tuning experiments (\secref{sec:finetuning}) were conducted on a single NVIDIA A10G GPU with 24\,GB of memory. We applied LoRA adapters with rank 16 and $\alpha{=}32$, which kept the number of trainable parameters small enough to fit within the memory budget of a single GPU even with full image inputs. Training times ranged from several minutes for the smallest sample budgets ($N{=}1$) to under one hour for the largest configurations, making the fine-tuning stage lightweight and easily reproducible.

\boldparagraph{Baseline evaluation.} Zero-shot evaluations of the baseline VLMs (\secref{sec:autoqa-results}) were performed using proprietary and open-weight models. Proprietary models were queried through cloud-based batch-inference APIs, completing a full evaluation run in approximately two hours per model. Open-weight models were served locally with \texttt{vLLM} on GPUs ranging from 48\,GB to 1128\,GB of memory depending on model size, and required over 36~hours of wall-clock time per run.

\boldparagraph{Orientation classification and VQA-based baselines.}
Inference with zero-shot orientation models (\secref{sec:orientation}) and VQA-based baselines (\secref{ssec:vqa-quality}) was performed locally on a single NVIDIA A10G GPU. Processing the full evaluation sets required about several hours for the orientation experiment and approximately 2 days for the VQA baselines.

\plotrejectedvcapessupp

\section{Fine-tuning on \datasetName: Implementation Details}\label{sec_ss:fewshot_finetuning_details}
In~\secref{sec:finetuning}, we fine-tune Qwen3-VL-8B-Instruct that supports multi-image inputs and instruction following. We apply LoRA to the query and value projection matrices (\texttt{q\_proj}, \texttt{v\_proj}) of the attention layers, with rank $r{=}16$, scaling factor $\alpha{=}32$ (effective scaling $\alpha/r = 2$), and dropout $p{=}0.05$. No bias terms are adapted. This configuration introduces approximately 0.1\% additional trainable parameters relative to the frozen base model, enabling efficient adaptation on a single GPU.

\boldparagraph{Training procedure.} We train for 5 epochs using the AdamW optimizer~\cite{loshchilov2019adamw} with a peak learning rate of $2 \times 10^{-4}$, a warmup ratio of 3\%, and cosine decay. The effective batch size is 4 (per-device batch size of 1 with 4 gradient accumulation steps). We apply gradient clipping with a maximum norm of 0.3 and use \texttt{bfloat16} mixed precision throughout. Gradient checkpointing is enabled to reduce memory consumption. For configurations with more than two training samples, we hold out 20\% of the data as a validation set and select the checkpoint with the lowest evaluation loss; otherwise, we use the final checkpoint.~\figref{fig:finetune_dynamics} reports the training loss progression for the $N{=}5$ configuration (11 training examples, 15 gradient steps over 5 epochs). The loss decreases monotonically from 4.93 at step~1 to 0.5 at step~110, with mean token accuracy improving from 42.0\% to 92.0\%.

\section{Societal Impact}\label{sec_ss:soc_impact}

This work introduces a benchmark for automated quality assessment of generative image editing, with a focus on preserving fine-grained visual fidelity in e-commerce product imagery.
We discuss both the potential benefits and risks associated with this line of research.

\boldparagraph{Positive impacts.} Reliable automated quality assessment can reduce the need for large-scale manual annotation, lowering costs and improving turnaround times for content moderation in e-commerce platforms. By detecting subtle visual inconsistencies, such as altered geometry, incorrect pose, or hallucinated elements, AutoQA systems could help maintain product representation accuracy, supporting informed purchasing decisions and reducing return rates driven by misleading imagery. More broadly, our benchmark highlights critical limitations of current vision-language models in fine-grained visual understanding, which may guide future model development toward more robust and trustworthy visual reasoning.

\boldparagraph{Potential risks.} The dataset and methods presented here could lower the barrier to deploying generative editing pipelines at scale, potentially enabling the creation of misleading or manipulated product imagery if quality controls are circumvented rather than enforced.
While our work is designed to \emph{detect} such failures, the same understanding of model blind spots could, in principle, be used to craft adversarial edits that evade automated checks.

Additionally, increased automation of visual quality assessment may reduce demand for human annotators and content reviewers, with implications for workers currently employed in these roles.
We note, however, that our results demonstrate that fully automated assessment remains far from reliable, suggesting that human oversight will remain essential for the foreseeable future.

\boldparagraph{Data considerations.} \datasetNameReal is constructed from e-commerce catalog imagery and does not contain personally identifiable information.
\datasetNameSynthetic is derived from the publicly available Amazon Berkeley Objects dataset~\cite{abo_dataset}, which consists of 3D product assets. All human annotations were collected through structured annotation tasks with trained annotators. We do not foresee significant privacy or fairness concerns arising from the dataset content, as it is restricted to product imagery without depictions of individuals.

\annotationtemplate

\end{document}